\documentclass[conference]{IEEEtran}
\IEEEoverridecommandlockouts

\usepackage{cite}
\usepackage{amsmath,amssymb,amsfonts}
\usepackage[linesnumbered,ruled,noend]{algorithm2e}
\usepackage{graphicx}
\usepackage{textcomp}

\usepackage{booktabs}
\usepackage{multirow}
\usepackage{calligra}
\usepackage{pdfpages}
\usepackage{enumitem}
\usepackage{hhline}

\usepackage{booktabs}
\usepackage{subcaption}
\usepackage{colortbl}
\usepackage{balance}
\usepackage{caption}
\usepackage{hyperref}
\usepackage{amstext}
\usepackage[misc]{ifsym}
\usepackage{pifont}

\usepackage{makecell}
\newcolumntype{C}[1]{>{\centering\arraybackslash}m{#1}}

\usepackage[most]{tcolorbox}
\usepackage{dashrule}

\usepackage[capitalize]{cleveref}
\crefname{section}{Sec.}{Secs.}
\Crefname{section}{Section}{Sections}
\Crefname{table}{Table}{Tables}
\crefname{table}{Tab.}{Tabs.}

\newcommand{\ours}[0]{REAL\ }

\usepackage{xcolor}
\def\BibTeX{{\rm B\kern-.05em{\sc i\kern-.025em b}\kern-.08em
    T\kern-.1667em\lower.7ex\hbox{E}\kern-.125emX}}
\begin{document}

\title{
REAL: A Reasoning-Enhanced Graph Framework for Long-Term Memory Management of LLMs
}

\author{Keer Lu$^{\ddagger}$, Liwei Chen$^{\mathsection}$, Guoqing Jiang$^{\mathsection}$, Zhiheng Qin$^{\mathsection}$, 
Yunhuai Liu$^{\ddagger \dagger \P}$, Wentao Zhang$^{\dagger \P}$ \\
$^\ddagger$\textit{School of Computer Science, Peking University},  
$^\mathsection$\textit{Kuaishou Technology}, \\
$^\dagger$\textit{Center for Data Science, Academy for Advanced Interdisciplinary Studies, Peking University} \\ 
keer.lu@stu.pku.edu.cn, 
\{yunhuai.liu, wentao.zhang\}@pku.edu.cn
}

\maketitle

\begin{abstract}
Large Language Models (LLMs) are increasingly expected to interact with users over long time horizons. However, due to their finite context window, LLMs cannot retain all past interactions, making long‑term memory management essential for storing, updating, and retrieving historical information beyond the context limit.
Although recent memory systems attempt to address this issue by storing historical information externally, existing approaches suffer from three key limitations: 
flat text-based memory organizations fail to capture explicit relations among memories, structured memory systems often destructively overwrite evolving facts, and current retrieval mechanisms remain query-agnostic and passive when evidence is incomplete.

To address these challenges, we propose \textit{REAL}, a reasoning-enhanced graph framework for long-term memory management of LLMs. 
\textit{REAL} constructs long-term conversational memory as a temporal and confidence-aware directed property graph, where each atomic fact is represented with entities, relations, valid-time intervals, confidence scores, and exploration intent labels. During memory construction, \textit{REAL} adopts a non-destructive temporal update strategy that preserves parallel fact versions and their validity intervals, enabling faithful tracking of fact evolution. During retrieval, \textit{REAL} anchors query-relevant root entities, decouples their exploration intents, and performs semantic evaluator-guided hybrid beam search to extract compact memory subgraphs. It further incorporates counterfactual inference to repair unreliable retrieval states and recover missing memory evidence through implicit logical relations. Comprehensive experiments demonstrate that \textit{REAL} substantially improves long-term memory performance over flat-text, graph-based, and existing state-of-the-art memory management methods, achieving an average improvement of 22.72\%.
\end{abstract}

\begingroup\renewcommand\thefootnote{$\P$}
\footnotetext{Corresponding Authors.}
\endgroup

\begin{IEEEkeywords}
Long-Term Memory Management, Large Language Models, Graph Data, 
Hybrid Beam Search
\end{IEEEkeywords}

\section{Introduction}
\label{sec:intro}
Large Language Models (LLMs)~\cite{achiam2023gpt,yang2025qwen3,guo2025deepseek_r1} have demonstrated remarkable capabilities in long-duration tasks, interacting with users over long time horizons that span multiple sessions and hundreds of conversation turns~\cite{zhang2025memory_survey}. 
Such long‑term interactions highlight the importance of \textit{LLM memory}~\footnote{``Memory'' discussed in this paper is distinct from volatile hardware RAM.}, which is essential for retaining and retrieving relevant information over extended periods~\cite{hu2023chatdb,hatalis2023memory,kang2025memory}. 
Despite their potential, maintaining effective memory over prolonged periods remains a critical challenge. 
The context window of modern LLMs is severely limited, which typically ranges from 8K to 256K tokens. 
As tasks lengthen and the accumulated historical contents exceed the model's fixed context window, it faces inherent limitations that force migration of information to persistent storage~\cite{chhikara2025mem0}.

One straightforward solution is to increase the model's context window length~\cite{achiam2023gpt,team2024gemini}. 
However, these advances only postpone the inherent limitation without resolving it: 
\textbf{\textit{(1) Unbounded Context Accumulation}}: The infinitely accumulating dialogue history will inevitably exceed the finite context limits as interactions evolve over weeks or months. 
\textbf{\textit{(2) Attention Decay over Distance}}: 
Even if the past information is included in the same context window, since the model's attention mechanism degrades with distance~\cite{li2024mechanics}, it struggles to retrieve or utilize distant tokens effectively, especially when relevant details are scattered across many tokens~\cite{hatalis2023memory}. 
\textbf{\textit{(3) Thematic Discontinuity}}: 
More importantly, real‑world conversations often jump between topics, e.g., the user might mention a travel plan, then spend hours troubleshooting a software bug, before returning to discuss hotel recommendations for the journey. Key facts can become lost among vast amounts of unrelated tokens, which makes full‑context reasoning inefficient~\cite{levy-etal-2024-task}. 
Given these inherent limitations, simply enlarging the context window is neither sufficient nor sustainable.

To overcome these limitations, there is a need for LLMs to adopt memory systems that move beyond static context expansion, where the model could retain crucial information, integrate related concepts, and retrieve relevant details on demand. 
A more principled approach is to decouple the model's long-term memory from its finite working memory, i.e., the context window. This separation calls for an \textit{external memory repository} that persistently stores historical information outside the LLM, indexes it for efficient retrieval, and dynamically supplies relevant subsets back into the context window when needed~\cite{xu2025mem}. 
Such a repository acts as a dedicated \textit{data management system}: it extracts facts from the dialogue stream, organizes them into a queryable structure, supports incremental updates, and provides fast, accurate retrieval.

However, realizing such a memory repository is non‑trivial. 
Through analysis, we identified that the challenges in existing LLM long-term memory management strategies stem precisely from the lack of such a well-designed memory mechanism, and can be traced to the following three key aspects:

\begin{figure*}[t]
    \centering
    \includegraphics[width=\textwidth]{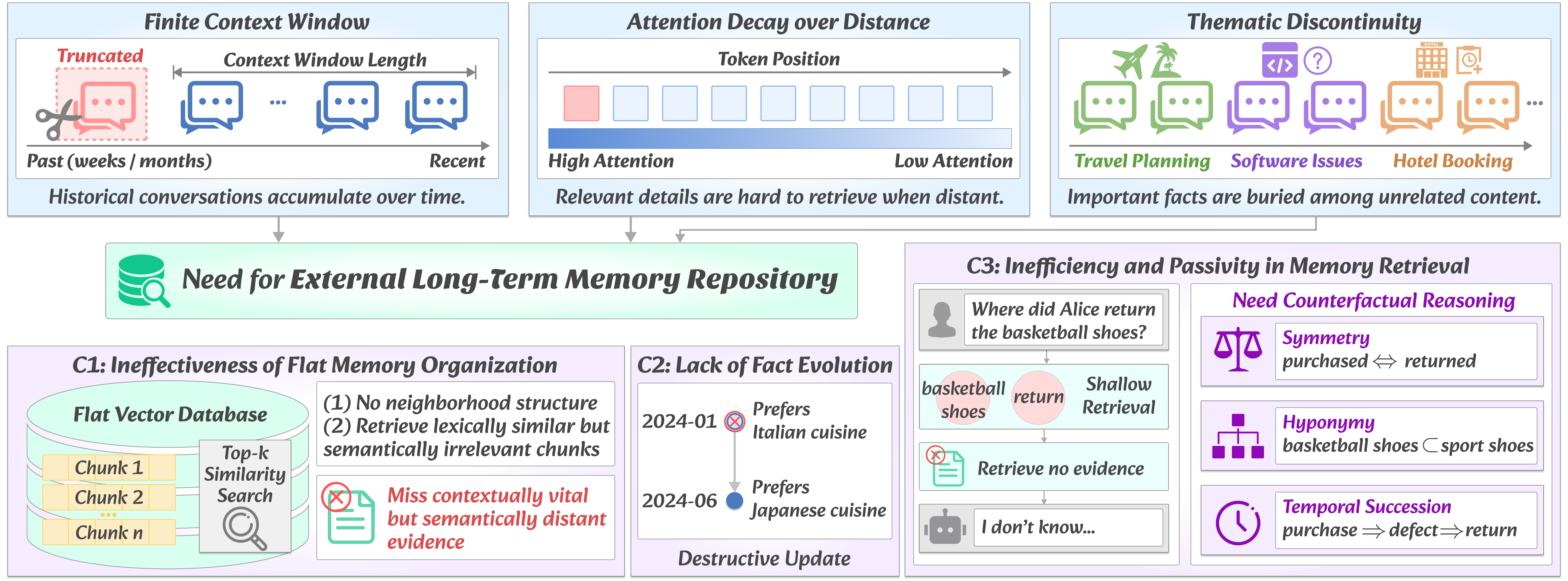}
    \caption{
    Illustration of existing challenges and design motivation. 
    Finite context windows, attention decay over long distances, and thematic discontinuity make it difficult for LLMs to preserve and retrieve historical information directly from the context window, motivating the need for an external memory repository. 
    However, existing memory systems suffer from: 
    \textbf{\textit{(C1)}} flat text-based organization lacks explicit memory links and multi-hop retrieval capability, 
    \textbf{\textit{(C2)}} destructive updates erase fact evolution, 
    and \textbf{\textit{(C3)}} query-agnostic and passive retrieval fails to recover missing evidence through counterfactual reasoning. 
    }
    \label{fig:introduction}
    \vspace{-3mm}
\end{figure*}

\textbf{\textit{C1: Ineffectiveness of Flat Memory Organization.}} 
Current memory repositories predominantly adopt flat text organization, where the accumulated history is segmented into text chunks, each independently embedded and stored in a vector database. 
Memory retrieval is then performed by nearest-neighbor search on the query embedding to return the top‑$k$ semantically similar chunks, which are then appended to the prompt~\cite{zhang2025memory_survey}. 
While this approach offers simplicity in implementation, it suffers from fundamental deficiencies that severely impede retrieval efficiency and contextual relevance. 
First, each text chunk is treated as an isolated semantic unit, and the repository maintains no explicit links between related chunks, making it impossible to navigate from a retrieved chunk to its logical neighbours without performing another similarity search. As a result, the system cannot perform path-based reasoning or multi-hop retrieval. 
Second, purely semantic similarity-based retrieval proves inadequate, which often retrieves lexically similar but semantically irrelevant chunks while missing contextually important evidence~\cite{hu2026does}.

\textbf{\textit{C2: Lack of Fact Evolution.}} 
Recognizing the limitations of flat text organizations, subsequent work has moved toward structured memory management. 
However, even when the memory is organized with structure, current LLM memory management systems adopt a destructive update paradigm, in which the newly acquired facts directly overwrite existing ones without preserving their evolutionary trajectories~\cite{chhikara2025mem0,xu2025mem}. 
Such replacement strategy fails to capture the temporal dynamics inherent in real-world knowledge, where the user preferences, plans, and relationships evolve continuously, and facts are refined over time~\cite{tan2026memotime}. For instance, when a user's preference changes from ``prefers Italian cuisine'' to ``prefers Japanese cuisine'', most studies simply replace the old fact with the new one, erasing the historical context that could inform future recommendations or explain behavioral shifts. 
This gap directly motivates our proposal: \textit{designing a structured memory management framework in which facts are annotated with explicit valid-time intervals, and conflicting updates are preserved as parallel versions instead of being destructively overwritten, thereby preserving the complete evolutionary history of every fact}.

\textbf{\textit{C3: Inefficiency and Passivity in Memory Retrieval.}} 
The retrieval mechanism of the memory repository itself persists as a critical bottleneck. 
Existing memory retrieval methods, whether based on flat vector similarity~\cite{pan2024survey}, tree‑based indexing~\cite{li2023constructing}, or graph traversal~\cite{rodriguez2012graph}, lack query‑aware exploration strategies. 
Conventional approaches, such as fixed‑step random walks~\cite{xia2019random}, or Personalized PageRank (PPR)~\cite{haveliwala2002topic}, execute the same retrieval pattern regardless of the query's specific intent. 
Tree‑based methods~\cite{li2023constructing} often rely on fixed‑depth traversal or static thresholds, missing deep or branching dependencies related to the query. 
Additionally, they are purely retrieval‑oriented and operate passively~\cite{ji2026memory}. 
When the retrieved evidence lacks sufficient coverage to answer the query, they simply return ``I don't know''. They cannot perform counterfactual reasoning, i.e., leveraging implicit logical relations such as symmetry (``likes'' $\Leftrightarrow$ ``dislikes'', ``purchased'' $\Leftrightarrow$ ``returned''), hyponymy (``basketball shoes'' $\subset$ ``sports shoes''), or temporal succession to infer new facts from the existing memory structure. 
Such passivity limits the model's ability to handle incomplete or ambiguous queries.

To address the challenges outlined above, we introduce \textbf{\textit{REAL}}, a \textbf{\textit{\underline{R}}}easoning-\textbf{\textit{\underline{E}}}nhanced gr\textbf{\textit{\underline{A}}}ph framework for \textbf{\textit{\underline{L}}}ong-term memory management of LLMs. 
For \textbf{\textit{C1}} and \textbf{\textit{C2}}, 
\ours organizes long-term conversational memory as a temporal, confidence-aware directed property graph, where each atomic fact is represented with entities, relations, valid-time intervals, confidence scores, and exploration intent labels. During memory construction, \ours incrementally updates the graph in a non-destructive manner, preserving their temporal evolution through parallel versions with explicit validity intervals. 
For \textbf{\textit{C3}} during memory retrieval, \ours first anchors query-relevant root entities and decouples their exploration intents, then performs semantic evaluator-guided hybrid beam search to extract a compact evidence subgraph. Each candidate traversal path is evaluated by query relevance, logical coherence, and entity-specific answer sufficiency, enabling the model to adaptively decide whether to stop or expand a beam. When normal expansion becomes unreliable, \ours further invokes a counterfactual inference mechanism to generate alternative traversal hypotheses and recover missing evidence from implicit relations.
In summary, our contributions are as follows:

\begin{itemize}[leftmargin=*]
    \item \underline{\textit{Multi-Attribute Memory Construction Mechanism.}} 
    We propose a multi-attribute memory construction mechanism that represents long-term conversational memory as a temporal, confidence-aware directed property graph, and adopt a non-destructive temporal update strategy that preserves parallel fact versions with explicit validity intervals 
    (for \textbf{\textit{C1}}, \textbf{\textit{C2}}).
    \item \underline{\textit{Reasoning-Enhanced Adaptive Retrieval Strategy.}} 
    We integrate mechanisms of exploration intent decoupling, semantic evaluator-guided hybrid beam search, and counterfactual inference to enable query-aware and robust memory evidence discovery during the memory retrieval phase 
    (for \textbf{\textit{C3}}).
    \item \underline{\textit{Performance and Effectiveness.}} 
    Comprehensive evaluations show that \textit{\ours} bolsters the model's long-term memory performances over existing state-of-the-art memory management methods with an average improvement of 22.72\%.
\end{itemize}

\section{Preliminary}
\label{sec:preliminary}

In this section, we introduce the background information and oundational concepts throughout the paper.

\subsection{Memory Mechanism of LLMs}
\label{subsec:llm_memory}

In real-world applications such as personal AI assistants~\cite{he2025llm_agent} or longitudinal medical consultations~\cite{wang2025medical_survey}, interaction histories often span months or even years. 
For these long-duration tasks, the memory mechanism of Large Language Models (LLMs) plays an extremely important role in determining how to accumulate knowledge, process historical experiences, retrieve relevant information to inform decisions, and so on~\cite{zhang2025memory_survey}.

\hypertarget{definition1}{\textbf{Definition \ref{sec:preliminary}.1 (Conversation Stream Modeling).}} 
Formally, we define a multi-turn conversation as a \textit{session} $U = (u_1, u_2, \dots, u_T)$, where each \textit{conversation turn} $u_t = \{c_t, \tau_t\}$ consists of the \textit{textual content} $c_t$, which is typically a user query or an LLM response, and an \textit{absolute timestamp} $\tau_t$ where $\tau_t \in \mathbb{R}_{\ge 0}$. 
Multiple sessions in the chronological order form a \textit{conversation stream} $\mathcal{X} = (U_1, U_2, \dots, U_N)$.

\hypertarget{definition2}{\textbf{Definition \ref{sec:preliminary}.2 (Memory Repository).}} 
Generally, the model maintains a \textit{memory repository} $\mathcal{M}$ to persistently store historical information extracted from the \textit{conversation stream} $\mathcal{X}$. 
$\mathcal{M}$ can adopt multiple data organization paradigms:
\begin{itemize}
    \item \textit{Flat-Text-based Structure}: Conversation stream $\mathcal{X}$ is split into fixed‑length text chunks and stored in a vector database. Retrieval is performed by nearest neighbor search to obtain the top‑$k$ semantically similar chunks.
    \item \textit{Graph‑based Structure}: These strategies incorporate graph databases for memory storage and retrieval processes, where the conversation stream $\mathcal{X}$ is represented as an entity‑relation graph to support structured path traversal and multi‑hop reasoning.
\end{itemize}

\subsection{Problem Formulation of LLM Memory Management}
\label{subsec:core_problems}

Given the conversation stream $\mathcal{X}$, the model should maintain a \textit{memory repository} $\mathcal{M}$ such that for any new user query $q$, the model can retrieve the most relevant subset $\mathcal{M}|_q$ from $\mathcal{M}$ and generate an accurate answer $a$. The core problems of memory management can be decomposed into two subproblems:

\begin{itemize}
    \item \textit{Memory Construction}: Given the accumulated conversation stream $\mathcal{X}$, the model transforms $\mathcal{X}$ into $\mathcal{M}$ via a memory construction process $\Phi_{\text{const}}$: 
    \begin{equation}
        \label{eq:mem_const}
        \mathcal{M} = \Phi_{\text{const}}(\mathcal{X})
    \end{equation}
    The central aspect of the research is how to extract, organize, and continuously update $\mathcal{M}$ from the streaming conversation input $\mathcal{X}$, such that $\mathcal{M}$ compactly represents historical information without losing critical details.
    \item \textit{Memory Retrieval}: For a new user query $q$, the model performs a memory retrieval process $\Phi_{\text{ret}}$:
    \begin{equation}
        \label{eq:mem_ret}
        \mathcal{M}|_q = \Phi_{\text{ret}}(q, \mathcal{M}) \subseteq \mathcal{M}
    \end{equation}
    where $\mathcal{M}|_q$ denotes the retrieved memory subset. The final answer $a$ is generated conditioned on the query $q$ and the memory subset $\mathcal{M}|_q$.
    The essence of the research is how to efficiently and accurately locate the subset of evidence $\mathcal{M}|_q$ from the large-scale memory repository $\mathcal{M}$ that best supports answering the query $q$.
\end{itemize}

\section{System Overview}
\label{sec:method}
\begin{figure*}[t]
    \centering
    \includegraphics[width=\textwidth]{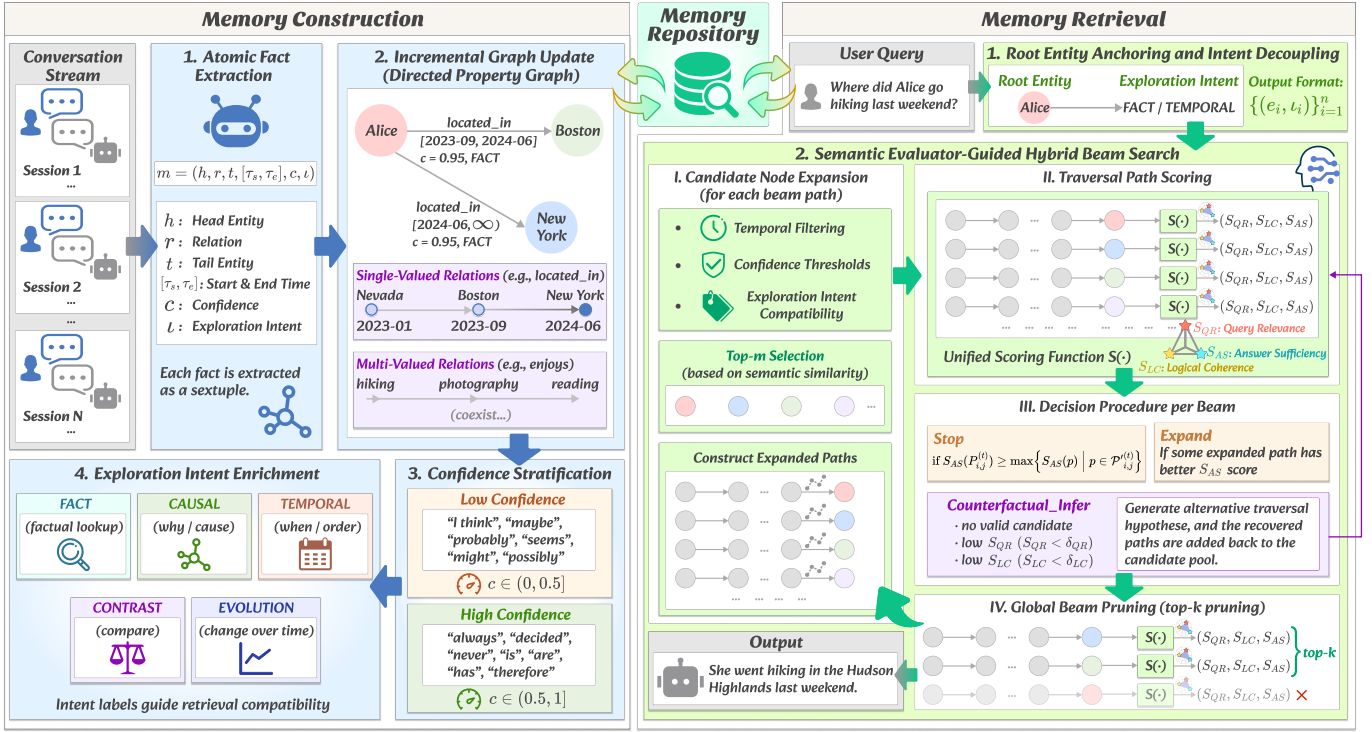}
    \caption{
    Overview of \ours pipeline. 
    During memory construction, conversation streams are converted into atomic facts in the form of \Cref{eq:atomic_fact} and incrementally stored in a temporal, confidence-aware directed property graph (\Cref{eq:mem_KG}). 
    During the memory retrieval phase, \ours anchors root entities from the query, decouples their exploration intents, and performs semantic evaluator-guided hybrid beam search to adaptively choose to stop, expand, or invoke counterfactual inference. 
    The retrieved paths are globally pruned and aggregated into a compact memory evidence subgraph for final answer generation.
    }
    \label{fig:pipeline}
    \vspace{-3mm}
\end{figure*}

In this section, we present the proposed \ours framework. 
As illustrated in \Cref{fig:pipeline}, \ours consists of the following core phases: memory construction, which incrementally builds a temporal and confidence-aware memory graph, and memory retrieval, which extracts a compact query-specific evidence subgraph for answer generation.
We also analyze the computational complexity and efficiency of REAL.

\subsection{Memory Construction}
\label{subsec:mem_construct}

The first phase in \ours framework is constructing the memory, which involves creating a multi-dimensional representation of entities and their relationships within a defined temporal context. This memory structure acts as the backbone for reasoning and subsequent memory retrieval. 
Here we represent the memory repository as a directed property graph: 
\begin{equation}
    \label{eq:mem_KG}
    \resizebox{1\linewidth}{!}{
    $
    \mathcal{M} = \{ (h, r, t, \tau, c, \iota) \mid h, t \in \mathcal{E}, r \in \mathcal{R}, \tau \subseteq [0,\infty), c \in (0, 1], \iota \in I \}
    $
    }
\end{equation}
where $\mathcal{E}$ denotes the set of all entities, e.g., users, objects, concepts, etc., $\mathcal{R}$ represents the set of all relations, and $(h, r, t)$ is a triple in which the head entity $h$ is connected to the tail entity $t$ via the relation $r$. 
Each edge carries a temporal interval $\tau = [\tau_s, \tau_e]\subseteq [0,\infty)$, along with a confidence score $c\in (0, 1]$ and an exploration intent label $\iota$.

\textbf{\textit{Atomic Fact Extraction.}} Given an incoming conversation stream $\mathcal{X}$, we segment it into sessions and then into individual turns. For each turn, we call an LLM to extract a set of atomic facts. Each fact is output in the form of a sextuple: 
\begin{equation}
    m = (h, r, t, [\tau_s, \tau_e], c, \iota)
    \label{eq:atomic_fact}
\end{equation}
where $\tau_s$ is the utterance timestamp or the explicit time mentioned in the utterance, and $\tau_e$ is initially set to $\infty$, which means ``still valid''. The LLM is prompted to assign confidence scores and exploration intents based on linguistic cues. 
For instance, statements containing hedging modal verbs, e.g., ``might'' and ``perhaps'', receive lower confidence, whereas definitive declarative statements are assigned higher confidence. 
Moreover, words like ``because'' trigger the exploration intent of \texttt{CAUSAL}, while ``before/after'' trigger \texttt{TEMPORAL}. 
We then detail how these attributes, incluing the temporal interval $[\tau_s,\tau_e]$, confidence score $c$, and exploration intent $\iota$, are determined and maintained during memory construction:

\textbf{Incremental Graph Update with Non‑Destructive Temporal Evolution.} 
When a new atomic fact $m_{\text{new}}=(h, r, t,\tau_{\text{new}},c_{\text{new}})$ arrives, we first retrieve any existing fact $m_{\text{old}}=(h, r, t',\tau_{\text{old}},c_{\text{old}})$ with the same head entity and relation but possibly different tail entity. The update behavior is determined by the \textit{cardinality type} of relation $r$: 

\begin{itemize}
    \item \textit{Single‑Valued Relations}: At any valid time interval, at most one tail entity can be true, e.g., ``current\_living\_city''. 
    If there exists $m_{\text{old}}$ with the same tail entity $t' = t$, we then merge the intervals $\tau_{\text{new}} = [\min(\tau_{s,\text{new}}, \tau_{s,\text{old}}), \infty)$. 
    Otherwise, if there exists $m_{\text{old}}$ with a different tail entity $t' \neq t$, and the intervals overlap, it indicates that a \textit{conflicting fact} occurs. In this case, we do not overwrite. Instead, we close the old interval by setting its $\tau_e$ to $\tau_{s,\text{new}}$, when the time the new fact $m_{\text{new}}$ becomes true, and insert the new fact with its own interval. Parallel edges with non‑overlapping intervals thus record the complete evolution of the attribute.
    \item \textit{Multi‑Valued Relations}: Multiple values can coexist even with overlapping intervals, e.g., ``likes\_food'', ``friend\_of''. 
    We simply add $m_{\text{new}}$ as a new parallel edge, without closing any existing interval. If the exact same triple $(h, r, t)$ already exists with an overlapping interval, we merge the intervals as in the same‑value case.
    \item \textit{Unknown Cardinality}: To avoid data loss, we regard the cardinality type of relation $r$ default to \textit{multi‑valued relations} in this case, while the system may later learn cardinality constraints from observed patterns. 
\end{itemize}

This cardinality‑aware temporal update strategy ensures that we neither mistakenly erase valid parallel facts (e.g., two liked foods) nor fail to capture genuine attribute evolution (e.g., a living city change). It forms the foundation for faithful historical backtracking in long‑term conversation memory.

\textbf{Confidence Stratification and Upgrade.}  
To prevent unconfirmed statements during memory retrieval, we introduce the confidence stratification strategy. 
Each atomic fact stored in the memory graph $\mathcal{M}$ is associated with a confidence score $c \in (0,1]$, which is initially assigned by the LLM during fact extraction according to linguistic reliability cues. 
Specifically, statements containing hedging expressions or modal verbs, such as ``might'', ``perhaps'', and ``maybe'', are assigned lower confidence scores, whereas assertive declarative statements, such as those containing ``is'', ``was'', or ``always'', are assigned higher confidence scores. 
During the memory retrieval phase, only facts whose current confidence satisfies $c \geq \theta_{\text{stable}}$ are considered as core memory evidence for beam expansion. 
Facts below this threshold are excluded from normal retrieval to reduce noise, unless the query explicitly requests uncertain or hypothetical information, e.g., the query contains uncertainty keywords such as ``guess'', ``maybe'', ``speculate''. 
When subsequent dialogue turns provide additional confirmation for a low-confidence fact, its confidence score can be upgraded accordingly, allowing previously uncertain memories to become eligible for future retrieval.

\textbf{Exploration Intent Enrichment.}
During memory construction phase, we further attach an exploration intent label 
$\iota \in \{$\texttt{FACT}, \texttt{CAUSAL}, \texttt{TEMPORAL}, \texttt{CONTRAST}, \texttt{EVOLUTION}$\}$ 
to each extracted atomic fact, which describes what type of future retrieval or reasoning the memory fact may support. 
For example, 
the sentence ``I used to prefer Italian food, but recently I choose Japanese food more often'' yields facts with an \texttt{EVOLUTION} intent, because the memory describes a change in user preference over time. 
A conversation record such as ``I stopped using the fitness app because it kept sending too many notifications'' is labeled a \texttt{CAUSAL} intent, since the fact supports future why/how reasoning. 
Similarly, utterances involving explicit time markers, e.g., ``before moving to Seattle'', are assigned \texttt{TEMPORAL} intent, while utterances comparing alternatives or expressing conflicts are assigned \texttt{CONTRAST} intent.
The stored intent label $\iota$ is later used during memory retrieval. 
When a user query is issued, the root entity anchoring module extracts the query-side exploration intent, and the beam search prioritizes memory edges whose stored exploration intent labels are compatible with the query intent.

\subsection{Memory Retrieval}
\label{subsec:mem_retrieval}

\begin{algorithm}[t]
\caption{Semantic Evaluator-Guided Hybrid Beam Search for Memory Retrieval (\Cref{subsec:mem_retrieval})}
\label{alg:semantic_hybrid_beam}
\small
\SetAlgoLined
\SetKwProg{Fn}{Function}{}{end}
\SetKwInput{KwData}{Input}
\SetKwInput{KwParameter}{Parameter}
\SetKwInput{KwDefine}{Define}
\SetKwInput{KwResult}{Ouptut}
\KwData{\mbox{User query $q$, Memory repository $\mathcal{M}$, Beam width $k$,} \mbox{Maximum search depth $D_{\text{max}}$, 
Stable confidence threshold} \mbox{$\theta_{\text{stable}}$, 
Anchored entities with exploration intents 
$\{(e_i, \iota_i)\}_{i=1}^{n}$,} 
\mbox{Relevance threshold $\delta_{QR}$, Logical coherence threshold $\delta_{LC}$\;} 
}
\KwParameter{\mbox{Memory beam $\mathcal{B}$, Semantic evaluation score $S(\cdot)$\;}}
\KwOut{
Retrieved memory subset $\mathcal{M}_q$\;
}

$\mathcal{P}_{final} \leftarrow \emptyset$\;

\For{each $(e_i,\iota_i) \in \{(e_i, \iota_i)\}_{i=1}^{n}$}{
    Initialize beam set:
    $\mathcal{B}_{i}^{(0)} \leftarrow \{(e_i,\iota_i)\}$\;
    
    \For{$t=0$ \KwTo $D_{\text{max}}-1$}{
        $\mathcal{B}_{cand(i)}^{(t)} \leftarrow \emptyset$\;
        
        \For{each traversal memory path $P_{i,j}^{(t)} \in \mathcal{B}_{i}^{(t)}$}{
            \tcp{Candidate Node Expansion}
            \mbox{Retrieve direct neighbors $\mathcal{N}(e_{i,j}^{(t)})$ of $e_{i,j}^{(t)}$ from $\mathcal{M}$\;}
            
            $\mathcal{N}_{raw(i,j)}^{(t)}, \mathcal{P'}_{i,j}^{(t)} \leftarrow \emptyset$\;
            
            \For{each $u \in \mathcal{N}(e_{i,j}^{(t)})$}{
                \If{\mbox{$\textsc{TimeValid}(u,q)$ 
                    \textbf{and} 
                    $c(e_{i,j}^{(t)},u) \geq \theta_{\text{stable}}$}
                    \textbf{and}
                    $\textsc{IntentCompatible}(\iota_u,\iota_{i,j}^{(t)})$
                }{
                    $\mathcal{N}_{raw(i,j)}^{(t)} 
                    \leftarrow 
                    \mathcal{N}_{raw(i,j)}^{(t)} \cup \{u\}$\;
                }
            }
            
            \mbox{$\mathcal{N}_{cand(i,j)}^{(t)} 
            \leftarrow 
            \textsc{TopM}(
                \mathcal{N}_{raw(i,j)}^{(t)}, 
                \textsc{Sim}(q, r(e_{i,j}^{(t)},u) \oplus u)
            )$\;}
            
            $\mathcal{P'}_{i,j}^{(t)} \leftarrow \big\{ P_{i,j}^{(t)} \oplus (u, \iota_u) \;\big|\; u \in \mathcal{N}_{\text{cand} (i,j)}^{(t)} \big\}$\;
            \tcp{Traversal Path Scoring}
            \For{each $p \in \mathcal{P'}_{i,j}^{(t)} \cup \{P_{i,j}^{(t)}\}$}{
                \mbox{Compute 
                $S(p) = (S_{QR}(p), S_{LC}(p), S_{AS}(p))$\;}
            }
            \tcp{Counterfactual Repair}
            \If{$\textsc{NeedCounterfactual}
                (\mathcal{P'}_{i,j}^{(t)},P_{i,j}^{(t)})$}
            {
                \mbox{$\mathcal{C}_{i,j}^{(t)} 
                \leftarrow 
                \textsc{CounterfactualInfer}
                (q,\mathcal{M},P_{i,j}^{(t)},\iota_{i,j}^{(t)})$\;}
                
                \mbox{$\mathcal{P}_{cf(i,j)}^{(t)}
                \leftarrow
                \textsc{ConstrainedTraverse}(\mathcal{M},\mathcal{C}_{i,j}^{(t)})$\;}
                $\mathcal{B}_{cand(i)}^{(t)}
                    \leftarrow
                    \mathcal{B}_{cand(i)}^{(t)} \cup \mathcal{P}_{cf(i,j)}^{(t)}$\;
            }
               \tcp{Stop-or-Expand Decision}
                \uIf{\mbox{$S_{AS}(P_{i,j}^{(t)}) \geq
                \max\{S_{AS}(p) \mid p \in \mathcal{P'}_{i,j}^{(t)}\}$}}{
                    $\mathcal{P}_{final}
                    \leftarrow
                    \mathcal{P}_{final} \cup \{P_{i,j}^{(t)}\}$\;
                }
                \Else{
                    \mbox{$\mathcal{P}_{better}
                    \leftarrow
                    \{ \, p \in \mathcal{P'}_{i,j}^{(t)}
                    \mid
                    S_{AS}(p) > S_{AS}(P_{i,j}^{(t)})\}$\;}
                    
                    $\mathcal{B}_{cand(i)}^{(t)}
                    \leftarrow
                    \mathcal{B}_{cand(i)}^{(t)} \cup \mathcal{P}_{better}$\;
                }
        }

        \If{$\mathcal{B}_{cand(i)}^{(t)} = \emptyset$}{
            \textbf{break}\;
        }
        \tcp{Global Beam Pruning}
        $\mathcal{B}_{i}^{(t+1)}
        \leftarrow
        \textsc{TopK}
        (
            \mathcal{B}_{cand(i)}^{(t)},
            S(P \mid q),
            k
        )$\;
        
        \If{all active beams have selected \texttt{Stop}}{
            \textbf{break}\;
        }
    }
}

$\mathcal{M}_q
\leftarrow
\textsc{AggregateEvidence}(\mathcal{P}_{final})$\;

\Return $\mathcal{M}_q$\;
\end{algorithm}

Given a user query $q$, the goal of the memory retreival stage is to produce a small memory subset $\mathcal{M}_q$ from the entire memory repository $\mathcal{M}$ that maximally supports answering $q$. The memory retrieval stage contains the following phases:

\subsubsection{Root Entity Anchoring and Intent Decoupling}
\label{subsubsec:entity_intent}

For each user query $q$, we prompt the LLM to automatically extract a set of root entities as entry points of the memory repository $\mathcal{M}$. Unlike conventional entity extraction that only identifies surface mentions, our root entity anchoring module further decouples the exploration intent of each entity. 
Specifically, the exploration intent is inferred from the semantic goal of the query, the entity role, the expected answer type, temporal expressions, dialogue history, and the relation schema of the memory repository. 
The output is in the form of $\{(e_i, \iota_i)\}_{i=1}^{n}$, 
where $e_i \in \mathcal{E}$ denotes the initial root entity for exploration, 
and $\iota_i \in I$ is the intent direction for that anchored entity.

\subsubsection{Semantic Evaluator-Guided Hybrid Beam Search}
\label{subsubsec:beam_search}

We formulate the expansion from each anchored entity as a hybrid beam search process that integrates reinforcement learning (RL) to learn optimal expansion actions. 
The model receives a user query $q$, a set of \textit{initial} candidate entities along with their exploration intents $\{(e_i, \iota_i)\}_{i=1}^{n}$, and aims to retrieve an optimal memory subset $\mathcal{M}_q$ within the entire memory repository $\mathcal{M}$ over a series of steps $t = 1, 2, ..., T$.

Here we set the beam width to be $k$. 
Start from each initial root entity $e_i$ in $\{(e_i, \iota_i)\}_{i=1}^{n}$, 
in the context of beam search, at each step $t$, the $j$-th beam $(j = 1, ..., k)$ independently maintains a traversal path $P_{i,j}^t$ from the root entity $e_i$ to the current node, which records both visited entities and their exploration intents, i.e., 
$P_{i,j}^{(t)}=(e_i, \iota_i, e_{i,j}^{(1)}, \iota_{i,j}^{(1)}, \ldots, e_{i,j}^{(t)}, \iota_{i,j}^{(t)})$.

The operational details are described below.

\textbf{I. \textit{Candidate Node Expansion.}} 
For each traversal memory path $P_{i,j}^{(t)}$ of the $j$-th beam $(j = 1, ..., k)$ at step $t$, we first retrieve the direct neighboring nodes of the current tail entity $e_{i,j}^{(t)}$ in $P_{i,j}^{(t)}$. 
To reduce the search space and ensure retrieval quality, we apply three cascading filters based on properties of the memory repository 
$\mathcal{M}$ constructed in \Cref{subsec:mem_construct}:
\begin{itemize}
    \item \textit{Temporal Filtering}: A neighbor is retained only if its valid time interval $\tau = [\tau_s, \tau_e]$ contains the query's reference time $\tau_q$ (if explicitly provided, e.g., ``last month''). 
    When no temporal constraint is given, it is default to the current dialogue time. Neighbors whose time intervals lie completely in the future or have expired before the query's temporal scope are excluded. 
    \item \textit{Confidence Thresholds}: Only neighbors with a confidence score $c \ge \theta_{\text{stable}}$ are considered for steady retrieval.
    Low‑confidence edges ($c < \theta_{\text{stable}}$) are excluded from the normal retrieval process, unless the query explicitly contains uncertainty markers (e.g., ``guess'', ``maybe''). 
    \item \textit{Exploration Intent Compatibility}: Each fact in $\mathcal{M}$ carries a pre‑assigned intent label $\iota_e.$ 
    The neighbor is kept only if $\iota_e$ is compatible with the current beam's exploration intent $\iota_{i,j}^{(t)}$. 
    For example, for a causal intent, only edges labeled \texttt{CAUSAL} (e.g., ``causes'', ``leads\_to'') are allowed, while for a temporal intent, edges labeled \texttt{TEMPORAL} (e.g., ``precedes'', ``follows'') are retained.
\end{itemize}

After applying the above filters, the remaining neighbor nodes form the candidate set $\mathcal{N}_{\text{raw} (i,j)}^{(t)}$. From $\mathcal{N}_{\text{raw} (i,j)}^{(t)}$, we further select the top‑$m$ candidates based on the semantic similarity score, e.g., cosine similarity between the query embedding and the concatenation of the edge's relation name and target entity in the atomic fact, denoted as $\mathcal{N}_{\text{cand} (i,j)}^{(t)}$.
\begin{equation}
    \mathcal{P'}_{i,j}^{(t)} = \big\{ P_{i,j}^{(t)} \oplus (u, \iota_u) \;\big|\; u \in \mathcal{N}_{\text{cand} (i,j)}^{(t)} \big\}
    \label{eq:expand_candidate}
\end{equation}
where $\oplus$ denotes path concatenation, and $\iota_u$ is the updated exploration intent after visiting $u$. The final candidate expanded paths $\mathcal{P'}_{i,j}^{(t)}$ is then passed to the semantic evaluator for scoring.

\textbf{II. \textit{Traversal Path Scoring.}} 
We employ the semantic evaluator to assess the current trsversal path $P_{i,j}^{(t)}$ and all possible expanded paths $\mathcal{P'}_{i,j}^{(t)}$ of the $j$-th beam $(j = 1, ..., k)$ at step $t$ under a unified scoring function. The scoring rubric considers: 
\begin{itemize}
    \item \textit{Query Relevance} ($S_{QR}$): The degree of semantic correlation between the trsversal memory path and the user's query. High relevance indicates that the path's content directly pertains to the core information need expressed in the query, while low relevance suggests off‑topic facts.
    For instance, a path describing ``user $\rightarrow$ likes $\rightarrow$ Italian cuisine'' would receive high relevance score for the query ``What is the user's favorite food?'', while a path about ``user $\rightarrow$ works $\rightarrow$ tech company'' would score low. 
    \item \textit{Logical Coherence} ($S_{LC}$): Whether the sequence of entities and relations in the trsversal memory path forms a plausible and consistent reasoning chain.
    For example, a path ``user $\rightarrow$ bought $\rightarrow$ red shoes'' followed by ``red shoes $\rightarrow$ made\_of $\rightarrow$ leather'' is coherent, whereas ``user $\rightarrow$ likes $\rightarrow$ red shoes'' immediately followed by ``red shoes $\rightarrow$ is $\rightarrow$ expensive'' without any bridging relation would be considered incoherent. 
    High logical coherence indicates that the path can serve as a reliable basis for reasoning, while low coherence suggests fragmentation or irrelevant jumps that may mislead the answer generation.
    \item \textit{Entity‑Specific Answer Sufficiency} ($S_{AS}$): 
    To what degree the current memory path $P_{i,j}^{(t)}$ contain enough information to answer the sub‑question that directly involves the root entity $e_i$ within the overall query. 
    Since each traversal memory path $P_{i,j}^{(t)}$ originates from a single root entity $e_i$ extracted from the query, it is not expected to fully resolve the entire multi‑entity or multi‑hop query. Instead, this metric evaluates whether the path already contains sufficient information to answer the part of the query that directly involves that root entity (e.g., the attribute value, the causal link, or the temporal event associated with this entity). A high score indicates that the sub‑question concerning this entity is satisfied, even if other parts of the query remain unanswered.
\end{itemize}

\textbf{III. \textit{Decision Procedure per Beam.}} 
After obtaining the unified scores for the current traversal path $P_{i,j}^{(t)}$ and its expanded candidates $\mathcal{P'}_{i,j}^{(t)}$ from the semantic evaluator, the next phase is to determine the operation for each beam by directly comparing the utility of stopping at the current path versus expanding to a new node. 
At each step $t$, for the $j$-th beam $(j = 1, ..., k)$ rooted at the initial entity $e_i$, its decision $a_{i,j}^{(t)}$ belongs to the decision space $\mathcal{A}$, which is defined as:
\begin{equation}
    \mathcal{A} = \{\texttt{Stop}, \texttt{Expand}, \texttt{Counterfactual\_Infer}\}
    \label{eq:action_space}
\end{equation}
The decision for the current beam is made by first comparing the \textit{entity‑specific answer sufficiency score} ($S_{AS}$) in a unified metric space. 
If the current path $P_{i,j}^{(t)}$ achieves a sufficiently higher score than all possible expanded paths $\mathcal{P'}_{i,j}^{(t)}$, the beam selects the \texttt{Stop} action:
\begin{equation}
    a_{i,j}^{(t)} = \texttt{Stop}, \quad \text{if } S_{AS}(P_{i,j}^{(t)}) \geq \max \left\{S_{AS}(p) \;\middle|\; p \in \mathcal{P'}_{i,j}^{(t)} \right\}
\end{equation}

Once the \texttt{Stop} decision is selected, the current memory path $P_{i,j}^{(t)}$ is added to the final evidence path pool. 
If one or more expanded paths obtain higher scores than the current path $P_{i,j}^{(t)}$, the \texttt{Expand} action is chosen. We insert those expanded paths whose entity-specific answer sufficiency score exceeds $S_{AS}(P_{i,j}^{(t)})$ into the global candidate path pool $\mathcal{B}_{cand (i)}^{(t)}$ of the initial entity $e_i$ for subsequent global top-$k$ beam pruning: 
\begin{equation}
    \mathcal{B}_{cand (i)}^{(t)}
    \leftarrow
    \mathcal{B}_{cand (i)}^{(t)}
    \cup
    \left\{ S_{AS}(p) > S_{AS}(P_{i,j}^{(t)}) \;\middle|\; p \in \mathcal{P'}_{i,j}^{(t)} \right\}
\end{equation}

The \texttt{Counterfactual\_Infer} action is triggered when the normal expansion process becomes unreliable. 
This occurs when one of the following conditions holds: 
\begin{itemize}
    \item No valid neighbor remains after filtering: $|\mathcal {N}_{\text{cand} (i,j)}^{(t)}| = 0$.
    \item The maximum query-neighbor similarity is below a threshold: 
    $\max \left\{S_{QR}(p) \;\middle|\; p \in \mathcal{P'}_{i,j}^{(t)} \;\right\} < \delta_{QR}$.
    \item All expanded candidates have low logical coherence quality: 
    $\max \left\{S_{LC}(p) \;\middle|\; p \in \mathcal{P'}_{i,j}^{(t)} \;\right\} < \delta_{LC}$.
\end{itemize}
where $\delta_{QR}$ denotes the minimum semantic similarity threshold between candidate neighbors and the query, and $\delta_{LC}$ denotes the minimum acceptable logical coherence quality.
\texttt{Counterfactual\_Infer} does not directly produce an answer, it repairs the retrieval process by generating alternative traversal hypotheses based on the rules described in \Cref{tab:counterfactual_inference_rules} instead. 
These hypotheses are converted into constrained graph traversals, and the recovered paths $\mathcal{P}_{cf(i,j)}^{(t)}$ are scored and merged back into the same global beam candidate pool $\mathcal{B}_{cand (i)}^{(t)}$. 
To prevent unbounded search overhead, we impose a counterfactual traversal budget 
$B_{cf(i,j)}^{(t)}=\max\{0, m-|\mathcal{N}_{rel(i,j)}^{(t)}|\}$ as the maximum number of counterfactually recovered paths retained for each triggered beam, 
where $m$ is the maximum candidate expansion size and $\mathcal{N}_{rel(i,j)}^{(t)}$ denotes the set of reliable normal candidates whose corresponding expanded paths satisfy the minimum query relevance and logical coherence requirements, i.e., 
$S_{QR}(p) \geq \delta_{QR}$ and $S_{LC}(p) \geq \delta_{LC}$. 
For the $j$-th beam rooted at $e_i$ at step $t$, the recovered path set satisfies $|\mathcal{P}_{cf(i,j)}^{(t)}| \leq B_{cf(i,j)}^{(t)}$. 
Only the top-$B_{cf(i,j)}^{(t)}$ recovered paths, ranked by the semantic evaluator, are merged back into the global beam candidate pool $\mathcal{B}_{cand(i)}^{(t)}$.

\begin{table}[tb]
    \centering
    \caption{Counterfactual inference rules.}
    \label{tab:counterfactual_inference_rules}
    \resizebox{\linewidth}{!}{
    \begin{tabular}{cc}
        \toprule
        \textbf{Category} & \textbf{Example} \\
        \midrule
        Symmetry & \texttt{likes(A,B)} $\rightarrow$ \texttt{interested\_in(A,B)} \\
        Hyponymy & \texttt{subClassOf(A,B)} $\wedge$ \texttt{likes(C,A)} $\rightarrow$ \texttt{likes(C,B)} \\
        Temporal Succession & \texttt{after(A,B)} $\wedge$ \texttt{happened(B)} $\rightarrow$ $\neg$\texttt{happened(A)} \\
        Causality & \texttt{causes(A,B)} $\wedge$ \texttt{true(A)} $\rightarrow$ \texttt{true(B)} \\
        Co‑Occurrence & \texttt{co\_occurs(A,B)} $\wedge$ \texttt{mentions(C,A)} $\rightarrow$ \texttt{may\_mention(C,B)} \\
        \bottomrule
    \end{tabular}
    }
\end{table}

\textbf{IV. \textit{Global Beam Pruning.}} 
After all beams have made their decisions at step $t$, all candidate paths are aggregated into the global candidate path pool $\mathcal{B}_{cand (i)}^{(t)}$, and only the top-$k$ paths are retained for the next iteration. 
The procedure continues until all beams select \texttt{Stop}, no valid candidate path remains, or the maximum search depth $D_{\text{max}}$ is reached.

\subsection{Answer Generation}
\label{subsec:answer_generation}

For answer generation, the retrieved memory subset $\mathcal{M}_q$ is serialized into a compact evidence context containing the selected paths, timestamps, and confidence scores. 
The serialization preserves both graph structure and metadata. 
For each retrieved path, we include its root entity, traversal sequence, relation labels, valid-time intervals, and confidence scores. 
A typical serialized evidence path is formatted as:
\begin{equation}
    e_0 \xrightarrow{r_1,\tau_1,c_1} e_1 
    \xrightarrow{r_2,\tau_2,c_2} \cdots 
    \xrightarrow{r_n,\tau_n,c_n} e_n
\end{equation}

Given the serialized memory context, the answer generator is instructed to produce a response to the user query $q$ based on the retrieved memory evidence.

\subsection{Complexity and Efficiency Analysis}
\label{subsec:complexity_analysis}

In this section, we analyze the computational complexity of REAL.  
Let $|\mathcal{M}|$ denote the number of atomic memory facts stored in the repository, $|\mathcal{E}|$ the number of entities, and $d$ the average out-degree of an entity in the memory graph. 
For a query $q$, $n$ is the number of anchored root entities, $k$ denotes the beam width, $D_{\text{max}}$ is the maximum search depth, and $m$ represents the number of top candidates retained after local neighbor filtering based on semantic similarity. 

\textbf{Memory Construction.}
For each incoming conversation turn, the model extracts a small set of atomic facts, each represented as 
$(h,r,t,[\tau_s,\tau_e],c,\iota)$ in the form of \Cref{eq:atomic_fact}. 
Given an index over the head entity and relation pair $(h,r)$, updating a new fact only requires retrieving existing facts with the same $(h,r)$ key rather than scanning the entire memory repository. 
Therefore, the graph update cost for each fact is $O(\deg_r(h))$, where $\deg_r(h)$ is the number of existing facts associated with the same head entity and relation. 
In practice, $\deg_r(h) \ll |\mathcal{M}|$, making the update process output-sensitive and scalable to continuously growing memory streams. 
The additional storage overhead of \ours is linear in the number of atomic facts, i.e., $O(|\mathcal{M}|)$, since each fact is stored once with a constant-size set of attributes described in \Cref{eq:atomic_fact}.

\textbf{Memory Retrieval.}
At query time, \ours starts from $n$ anchored root entities and performs hybrid beam search. 
At each step, each active beam visits the direct neighbors of its current tail entity, and the candidate filtering is bounded by the average degree $d$. 
Selecting the top-$m$ candidates results in $O(d\log m)$ time per beam step. 
Thus, the total cost of normal candidate expansion is $O\left(n \cdot k \cdot D_{\text{max}} \cdot d \log m \right)$. 
After candidate expansion, \ours evaluates the current path and its expanded paths using the semantic evaluator. 
Since each beam keeps at most $m$ expanded candidates, the number of path evaluations is bounded by $O\left(n \cdot k \cdot D_{\text{max}} \cdot (m+1)\right)$, where the additional one corresponds to the current path used for the stop-or-expand decision. 
In implementation, we batched these path evaluations into a single semantic-evaluation call per beam, which reduces the number of LLM calls from per-path scoring to per-beam scoring.

\textbf{Counterfactual Inference.}
This mechanism is only invoked when normal expansion becomes unreliable. 
Let $\rho$ denote the empirical triggering rate of counterfactual repair, and $\bar{B}_{cf}$ denote the average of the per-beam counterfactual traversal budget over all triggered beam steps. 
The additional retrieval cost introduced by counterfactual repair is 
$O\left(\rho \cdot n \cdot k \cdot D_{\text{max}} \cdot \bar{B}_{cf}\right)$. 
Since $\bar{B}_{cf}$ is budgeted and $\rho$ is typically much smaller than 1.0, counterfactual repair improves robustness without bringing unbounded graph exploration.

\textbf{Global Beam Pruning.}
At each beam step $t$, the candidate pool for each root entity $e_i$ contains the reliable normal candidates and the counterfactually recovered candidates from all active beams. 
Since the counterfactual traversal budget is defined as 
$B_{cf(i,j)}^{(t)}=\max\{0,m-|\mathcal{N}_{rel(i,j)}^{(t)}|\}$, 
the recovered paths $\mathcal{P}_{cf(i,j)}^{(t)}$ only fill the remaining reliable candidate slots of the current beam. 
Therefore, for each beam, the number of candidates inserted into the global pool is bounded by $|\mathcal{N}_{rel(i,j)}^{(t)}| + |\mathcal{P}_{cf(i,j)}^{(t)}| \leq |\mathcal{N}_{rel(i,j)}^{(t)}| + B_{cf(i,j)}^{(t)} \leq m$, which means the candidate pool for each root entity $e_i$ contains at most $k \cdot m$ paths at step $t$. 
Keeping the top-$k$ candidates can be implemented with a bounded heap, requiring $O(km\log k)$ per root entity per step.
Therefore, the overall retrieval complexity is $O\left(n \cdot D_{\text{max}} \left[ k d \log m + k(m+1) + \rho k \bar{B}_{cf} + km\log k \right] \right)$. 
Under this setting, since $k$, $D_{\text{max}}$, and $m$ are controlled hyperparameters, the query-time retrieval cost is primarily determined by the local neighborhood size, i.e., the average out-degree $d$ of an entity in the memory graph, where $d \ll |\mathcal{M}|$.

\section{Experiments and Results}
\label{sec:experiment}

\subsection{Experimental Setup}

\textbf{Benchmarks.} 
\textit{Dialogue-style memory benchmarks} simulate real-world multi-session conversations, which directly evaluates long-duration conversational memory, personalization, temporal updates, and multi-session reasoning. 
For this scenario, we have selected the following datasets:
(1) \textbf{\textit{LoCoMo}}~\cite{maharana2024evaluating}: 
A benchmark designed to evaluate whether language models can maintain and retrieve information from extended multi-session dialogues, where each conversation contains an average of approximately 300 turns and 9K tokens. 
(2) \textbf{\textit{LongMemEval}}~\cite{wulongmemeval}: 
A comprehensive benchmark that evaluates five core memory abilities: information extraction, multi-session reasoning, temporal reasoning, knowledge updates, and abstention.
(3) \textbf{\textit{PersonaMem}}~\cite{jiang2025know}: 
The benchmark encompasses curated user profiles and simulated user-LLM interaction histories, with each history containing up to 60 sessions across diverse real-world personalization tasks.
\textit{Long-form, document-style benchmarks} provide controlled environments for evaluating evidence retrieval, multi-hop path reasoning, and document-level knowledge composition. 
Given this setting, we adopt the following datasets:
(1) \textbf{\textit{HotpotQA}}~\cite{yang2018hotpotqa}: 
A widely used multi-hop question answering benchmark built from Wikipedia, which contains 113K question-answer pairs that require models to locate and reason over multiple supporting documents. 
(2) \textbf{\textit{2WikiMultihopQA}}~\cite{ho2020constructing}: 
This benchmark combines structured and unstructured data, exploiting Wikidata-style structured relations together with Wikipedia textual evidence, which provides more explicit and comprehensive reasoning supervision on long-term memory tasks. 
(3) \textbf{\textit{MuSiQue}}~\cite{trivedi2022musique}: 
It is a challenging benchmark comprising approximately 25K questions that require 2–4 reasoning hops, with explicitly connected steps where one hop often depends on the answer to a previous one.

\textbf{Baselines.} 
We compare \ours against several representative baselines that cover full historical context, flat text-based memory strategies, and graph-based memory approaches. 
\begin{itemize}[leftmargin=*]
    \item \textit{Full Context}: 
    This strategy directly concatenates the entire conversation history into the model's input context window.
    \item \textit{Flat Memory Chunk}: In this setting, the dialogue history is segmented into flat text chunks and stored in a vector database. At inference time, the top-$k$ most similar chunks are retrieved according to cosine similarity, and then concatenated with the query as context for answer generation.
    \item \textit{MemGPT}~\cite{packer2023memgpt}: 
    Inspired by operating systems, this memory-augmented framework introduces virtual context management to organize memory into different tiers and move information between them when needed.
    \item \textit{Mem0}~\cite{chhikara2025mem0}: 
    This memory-centric framework dynamically extracts and consolidates information from ongoing conversations to support long-term conversational coherence. It represents memories primarily as textual memory entries and retrieves them through semantic matching. 
    \item \textit{Vanilla Memory Graph}: 
    This baseline extracts facts as entity-relation triples. During retrieval, it starts from entities mentioned in the query and performs fixed-depth graph traversal according to simple graph proximity.
    \item \textit{Mem0$^g$}~\cite{chhikara2025mem0}: 
    It denotes the graph-memory variant of \textit{Mem0}. Different from the \textit{vanilla memory graph} strategy, its update phase employs conflict detection and resolution mechanisms when integrating new information into the existing graph.
    \item \textit{A-MEM}~\cite{xu2025mem}: 
    Drawing on the Zettelkasten principle~\cite{ahrens2022take}, it dynamically organizes memory into interconnected graphs. When a new memory is added, the system constructs a structured memory note with contextual descriptions, keywords, and tags, and then links it to related historical memories.
\end{itemize}

\textbf{Evaluation Metrics.} 
To assess the correctness of the final answer, we adopt two evaluation strategies. The first is Exact Match (\textit{EM}), which considers a prediction correct only when it exactly matches the ground‑truth answer. However, such exact match metric is too strict for our setting, since the result is described by natural language. Therefore, we also consider LLM-as-judge (\textit{LJ}) for automatic evaluation, where we use DeepSeek-V3~\cite{liu2024deepseek} to score answer correctness.

\begin{table*}[tb]
    \centering
    \caption{\label{tab:main_results} 
    Comparison of \ours with baselines on long-term memory-related benchmarks, reporting Exact Match (\textit{EM}, \%) and LLM-as-Judge (\textit{LJ}, \%) results. 
    The best results are highlighted in \textbf{bold}, and the second best results are \underline{underlined}.
    }
    \resizebox{\textwidth}{!}{
        \begin{tabular}{cc|cccccc|cccccc}
            \toprule
            \multirow{4}{*}{\textbf{Model}} & \multirow{4}{*}{\textbf{Method}} & \multicolumn{6}{c|}{\textit{Dialogue-Style Memory Benchmarks}} & \multicolumn{6}{c}{\textit{Long-Form, Document-Style Benchmarks}} \\
            \cmidrule(lr){3-8}
            \cmidrule(lr){9-14}
            & & \multicolumn{2}{c}{\textbf{LoCoMo}} & \multicolumn{2}{c}{\textbf{LongMemEval}} & \multicolumn{2}{c|}{\textbf{PersonaMem}} & \multicolumn{2}{c}{\textbf{HotpotQA}} & \multicolumn{2}{c}{\textbf{2WikiMultihopQA}} & \multicolumn{2}{c}{\textbf{MuSiQue}} \\
            & & \multicolumn{1}{c}{\textit{EM}} & \textit{LJ} & \textit{EM} & \textit{LJ} & \textit{EM} & \multicolumn{1}{c|}{\textit{LJ}} & \textit{EM} & \textit{LJ} & \textit{EM} & \textit{LJ} & \textit{EM} & \multicolumn{1}{c}{\textit{LJ}} \\
            
            \midrule
            \multicolumn{14}{c}{\textbf{\textit{Close-Sourced Frontier Models}}} \\
            \midrule
            \multirow{12}{*}{\textbf{DeepSeek-V3}}
            & Full Context 
            & 35.24 & 46.84 & 38.46 & 49.78 & 22.52 & 33.10 
            & 27.80 & 51.65 & 33.14 & 36.25 & 19.37 & 30.25 \\
            \cmidrule(lr){2-14}
            & \multicolumn{13}{l}{\textbf{\textit{Flat-Text-Based Memory Baselines}}} \\
            & Flat Memory Chunk 
            & 45.60 & 58.53 & 50.12 & 62.20 & 24.00 & 35.05 
            & 26.43 & 50.68 & 35.52 & 38.64 & 18.45 & 29.67 \\
            & MemGPT 
            & 48.32 & 61.28 & 53.46 & 65.84 & 25.87 & 37.56 
            & 28.15 & 52.37 & 34.46 & 36.65 & 20.59 & 32.36 \\
            & Mem0 
            & 52.65 & 66.67 & 58.73 & 71.28 & 30.14 & 42.46 
            & 27.65 & 51.79 & 37.75 & 39.68 & 23.57 & 34.61 \\
            \cmidrule(lr){2-14}
            & \multicolumn{13}{l}{\textbf{\textit{Graph-Based Memory Baselines}}} \\
            & Vanilla Memory Graph 
            & 49.43 & 62.64 & 54.20 & 66.08 & 27.36 & 38.98 
            & 29.28 & 54.05 & \underline{40.58} & \underline{43.87} & 22.25 & 33.65 \\
            & Mem0$^{g}$ 
            & 53.96 & 67.12 & \underline{60.80} & \underline{73.67} & 31.04 & 43.25 
            & 31.76 & 55.75 & 38.04 & 40.06 & 24.58 & 34.69 \\
            & A-MEM 
            & \underline{55.12} & \underline{68.80} 
            & 59.63 & 72.57
            & \underline{32.58} & \underline{45.45} 
            & \underline{32.80} & \underline{56.07} 
            & 39.95 & 42.33 
            & \underline{26.52} & \underline{35.90} \\
            \cmidrule(lr){2-14}
            & \cellcolor{blue!5}\textbf{\ours (ours)} & \multicolumn{1}{c}{\cellcolor{blue!5}\textbf{59.98}} & \cellcolor{blue!5}\textbf{73.76} & \cellcolor{blue!5}\textbf{65.92} & \cellcolor{blue!5}\textbf{78.58} & \cellcolor{blue!5}\textbf{37.19} & \multicolumn{1}{c|}{\cellcolor{blue!5}\textbf{49.21}} & \cellcolor{blue!5}\textbf{36.40} & \cellcolor{blue!5}\textbf{59.06} & \cellcolor{blue!5}\textbf{43.62} & \cellcolor{blue!5}\textbf{46.17} & \cellcolor{blue!5}\textbf{30.34} & \multicolumn{1}{c}{\cellcolor{blue!5}\textbf{41.73}} \\
            \midrule
            \multicolumn{14}{c}{\textbf{\textit{Open-Sourced Base / Instruct Models}}} \\
            \midrule
            \multirow{12}{*}{\textbf{Qwen3-32B}}
            & Full Context 
            & 31.06 & 43.28 & 35.14 & 46.52 & 19.20 & 30.16 
            & 24.35 & 45.80 & 29.93 & 32.45 & 12.49 & 24.54 \\
            \cmidrule(lr){2-14}
            & \multicolumn{13}{l}{\textbf{\textit{Flat-Text-Based Memory Baselines}}} \\
            & Flat Memory Chunk 
            & 42.36 & 55.10 & 47.00 & 59.38 & 20.85 & 32.14 
            & 25.25 & 47.10 & 30.49 & 33.67 & 18.25 & 27.45 \\
            & MemGPT 
            & 44.74 & 57.83 & 49.82 & 62.09 & 22.95 & 34.25 
            & 25.54 & 48.66 & 32.57 & 36.86 & 17.44 & 25.96 \\
            & Mem0 
            & 49.11 & 63.05 & 55.23 & 68.20 & 25.02 & 38.50 
            & 26.85 & 48.97 & 33.55 & 37.65 & 19.51 & 30.62 \\
            \cmidrule(lr){2-14}
            & \multicolumn{13}{l}{\textbf{\textit{Graph-Based Memory Baselines}}} \\
            & Vanilla Memory Graph 
            & 45.92 & 59.28 & 51.45 & 64.10 & 24.22 & 36.40 
            & 25.51 & 48.04 & 35.95 & 38.83 & 20.84 & 31.76 \\
            & Mem0$^{g}$ 
            & \underline{51.74} & \underline{66.30} & 56.48 & 69.35 & 28.54 & 40.42 
            & \underline{29.03} & \underline{51.37} & 35.05 & 37.80 & 21.54 & 32.09 \\
            & A-MEM 
            & 50.65 & 64.51
            & \underline{57.69} & \underline{70.45} 
            & \underline{31.34} & \underline{42.86} 
            & 28.25 & 50.63 
            & \underline{36.52} & \underline{39.16} 
            & \underline{22.85} & \underline{33.66} \\
            \cmidrule(lr){2-14}
            & \cellcolor{blue!5}\textbf{\ours (ours)} & \multicolumn{1}{c}{\cellcolor{blue!5}\textbf{54.88}} & \cellcolor{blue!5}\textbf{69.47} & \cellcolor{blue!5}\textbf{60.14} & \cellcolor{blue!5}\textbf{73.49} & \cellcolor{blue!5}\textbf{34.32} & \multicolumn{1}{c|}{\cellcolor{blue!5}\textbf{45.05}} & \cellcolor{blue!5}\textbf{30.56} & \cellcolor{blue!5}\textbf{53.71} & \cellcolor{blue!5}\textbf{39.07} & \cellcolor{blue!5}\textbf{42.79} & \cellcolor{blue!5}\textbf{25.57} & \multicolumn{1}{c}{\cellcolor{blue!5}\textbf{36.04}} \\
            
            \midrule
            \multirow{12}{*}{\textbf{\makecell{LLaMA-3.3-70B\\-Instruct}}}
            & Full Context 
            & 34.80 & 45.66 & 37.28 & 48.34 & 20.62 & 31.57 
            & 25.42 & 47.35 & 31.51 & 34.05 & 17.39 & 28.15 \\
            \cmidrule(lr){2-14}
            & \multicolumn{13}{l}{\textbf{\textit{Flat-Text-Based Memory Baselines}}} \\
            & Flat Memory Chunk 
            & 43.87 & 56.48 & 48.65 & 60.05 & 22.40 & 33.68 
            & 25.54 & 47.65 & 30.56 & 32.63 & 19.46 & 30.95 \\
            & MemGPT 
            & 46.16 & 59.40 & 51.28 & 63.44 & 24.20 & 35.73 
            & 27.57 & 49.67 & 32.15 & 34.62 & 18.46 & 29.61 \\
            & Mem0 
            & 50.67 & 64.85 & 56.58 & 69.91 & 28.42 & 41.06 
            & 28.86 & 50.73 & 35.55 & 38.16 & 21.75 & 32.65 \\
            \cmidrule(lr){2-14}
            & \multicolumn{13}{l}{\textbf{\textit{Graph-Based Memory Baselines}}} \\
            & Vanilla Memory Graph 
            & 47.40 & 61.02 & 52.80 & 65.46 & 25.74 & 37.52 
            & 27.65 & 50.69 & 33.52 & 35.67 & \underline{24.96} & \underline{35.62} \\
            & Mem0$^{g}$ 
            & 52.12 & 66.35 & 57.88 & 70.97 & \underline{31.19} & \underline{43.25} 
            & 30.62 & 52.27 & \underline{41.58} & \underline{45.79} & 23.25 & 32.98 \\
            & A-MEM 
            & \underline{55.38} & \underline{70.72} 
            & \underline{59.08} & \underline{72.17} 
            & 30.06 & 42.80
            & \underline{31.16} & \underline{53.70} 
            & 38.75 & 41.65 
            & 24.55 & 33.86 \\
            \cmidrule(lr){2-14}
            & \cellcolor{blue!5}\textbf{\ours (ours)} & \multicolumn{1}{c}{\cellcolor{blue!5}\textbf{58.69}} & \cellcolor{blue!5}\textbf{72.83} & \cellcolor{blue!5}\textbf{62.78} & \cellcolor{blue!5}\textbf{76.05} & \cellcolor{blue!5}\textbf{35.98} & \multicolumn{1}{c|}{\cellcolor{blue!5}\textbf{48.28}} & \cellcolor{blue!5}\textbf{33.64} & \cellcolor{blue!5}\textbf{56.77} & \cellcolor{blue!5}\textbf{45.65} & \cellcolor{blue!5}\textbf{48.70} & \cellcolor{blue!5}\textbf{28.59} & \multicolumn{1}{c}{\cellcolor{blue!5}\textbf{39.18}} \\

            \bottomrule
        \end{tabular}
    }
\vspace{-0.1cm}
\end{table*}

\textbf{Models and Implementation.} 
To ensure fair evaluation, we use \textit{the same backbone model} throughout the end-to-end (E2E) experiments, isolating performance gains as attributable to the proposed memory framework, not to differences in model capability.
We utilize DeepSeek-V3~\cite{liu2024deepseek} as the close-sourced frontier models, while using Qwen3-32B~\cite{yang2025qwen3} and LLaMA3.3-70B-Instruct~\cite{dubey2024llama} as the open-sourced backbone models in our main experiments. 
For candidate pre-ranking, i.e., selecting $\mathcal{N}_{\text{cand} (i,j)}^{(t)}$ from $\mathcal{N}_{\text{raw} (i,j)}^{(t)}$, we employed the BGE-m3 model~\cite{chen2024bge} as our embedding model. 
When conducting evaluations, we used a single-node machine equipped with 8 NVIDIA H800 GPUs. Depending on the CUDA memory requirements and computational demands of tasks, the assessment process for each task utilized between 1 to 8 GPUs.

\textbf{Hyper-Parameters Setting.} 
During the memory retrieval phase, the beam width $k$, maximum search depth $D_{\text{max}}$, maximum number of candidate neighbors per expansion $m$, stable confidence threshold $\theta_{\text{stable}}$, query relevance threshold $\delta_{QR}$, and logical coherence threshold $\delta_{LC}$ were adjusted to optimize the balance between retrieval accuracy and computational cost. 
As analyzed in \Cref{subsubsec:search_hyper}, we have performed experiments under a range of hyperparameter settings with the Qwen3-32B model, and selected the optimal settings for hyperparameters in our \ours framework. 
Overall, the default configuration used in our main experiments is $k=5$, $D_{\text{max}}=3$, $m=5$, $\theta_{\text{stable}}=0.8$, $\delta_{QR}=0.5$, and $\delta_{LC}=0.5$.

\subsection{Main Results}

The main results of baselines and \ours are demonstrated in \Cref{tab:main_results}.
and we summarize the observations below.

\textbf{\ours is effective across different models.} 
Experimental results in \Cref{tab:main_results} show that \ours consistently outperforms other baseline methods across all backbone models in terms of long-term memory management.
Compared with the \textit{full context}, which directly feeds the full context into the model, \ours enhances average downstream task performances by 49.33\%, 
which suggests that simply exposing the model to the entire history is not sufficient for reliable long-term memory reasoning. 
As opposed to \textit{flat-text-based memory methods}, including \textit{Flat Memory Chunk}, \textit{MemGPT}, and \textit{Mem0}, our method yields a 23.85\% gain, 
underscoring its superiority over approaches that rely on isolated text chunks or plain textual memory units. 
Moreover, models enhanced with \ours
and it exceeds the average performance of \textit{graph-based memory baselines}, such as \textit{vanilla memory graph}, \textit{Mem0$^{g}$}, and \textit{A-MEM}, by 12.71\%, 
further demonstrating that merely converting memory into a graph is not enough, where temporal and reasoning‑aware memory mechanisms are essential.

\begin{table*}[tb]
    \centering
    \caption{\label{tab:component_ablation}
    Ablations on the impact of each component using the \textbf{Qwen3-32B} model, 
    reporting Exact Match (\textit{EM}, \%) and LLM-as-Judge (\textit{LJ}, \%) results. 
    ``\textbf{\textit{w/o}}'': \textit{removing} the component from the original \ours while keeping the remaining settings unchanged.
    The best and second best scores are in \textbf{bold} and \underline{underlined}.
    }
    \resizebox{\textwidth}{!}{
        \begin{tabular}{cccccccccccccc}
            \toprule
            \multirow{4}{*}{\textbf{Method}} 
            & \multicolumn{6}{|c|}{\textit{Dialogue-Style Memory Benchmarks}} 
            & \multicolumn{6}{c|}{\textit{Long-Form, Document-Style Benchmarks}} 
            & \multirow{3}{*}{\textbf{Avg.}} \\
            \cmidrule(lr){2-7}
            \cmidrule(lr){8-13}
            & \multicolumn{2}{|c}{\textbf{LoCoMo}} 
            & \multicolumn{2}{c}{\textbf{LongMemEval}} 
            & \multicolumn{2}{c|}{\textbf{PersonaMem}} 
            & \multicolumn{2}{c}{\textbf{HotpotQA}} 
            & \multicolumn{2}{c}{\textbf{2WikiMultihopQA}} 
            & \multicolumn{2}{c|}{\textbf{MuSiQue}} 
            & \\
            & \multicolumn{1}{|c}{\textit{EM}} 
            & \textit{LJ} 
            & \textit{EM} 
            & \textit{LJ} 
            & \textit{EM} 
            & \multicolumn{1}{c|}{\textit{LJ}} 
            & \textit{EM} 
            & \textit{LJ} 
            & \textit{EM} 
            & \textit{LJ} 
            & \textit{EM} 
            & \multicolumn{1}{c|}{\textit{LJ}} 
            & \textit{LJ} \\
            \midrule 

            \textbf{\ours} 
            & \multicolumn{1}{|c}{\textbf{54.88}} 
            & \textbf{69.47} 
            & \textbf{60.14} 
            & \textbf{73.49} 
            & \textbf{34.32} 
            & \multicolumn{1}{c|}{\textbf{45.05}} 
            & \textbf{30.56} 
            & \textbf{53.71} 
            & \textbf{39.07} 
            & \textbf{42.79} 
            & \textbf{25.57} 
            & \multicolumn{1}{c|}{\textbf{36.04}} 
            & \textbf{53.43} \\
            \cmidrule(lr){1-14}

            \multicolumn{14}{l}{\cellcolor[HTML]{EFEFEF}\textbf{\textit{Memory Construction Phase}}} \\

            \textit{w/o} Temporal Interval 
            & \multicolumn{1}{|c}{53.42} 
            & 68.27 
            & 58.05 
            & 71.61 
            & 32.55 
            & \multicolumn{1}{c|}{43.23} 
            & \underline{30.42} 
            & \underline{53.50} 
            & 38.55 
            & 42.40 
            & 25.34 
            & \multicolumn{1}{c|}{35.86} 
            & 52.48$^{\textcolor{red}{\downarrow0.95}}$ \\

            \textit{w/o} Confidence Stratification 
            & \multicolumn{1}{|c}{\underline{54.44}} 
            & \underline{68.85} 
            & 58.59 
            & 72.15 
            & \underline{33.96} 
            & \multicolumn{1}{c|}{\underline{44.57}} 
            & 30.26 
            & 53.13
            & \underline{38.98} 
            & \underline{42.55} 
            & \underline{25.58} 
            & \multicolumn{1}{c|}{\underline{35.95}} 
            & \underline{52.87}$^{\textcolor{red}{\downarrow0.56}}$ \\

            \textit{w/o} Exploration Intent 
            & \multicolumn{1}{|c}{53.20} 
            & 67.62 
            & \underline{59.45} 
            & \underline{72.64}
            & 33.15 
            & \multicolumn{1}{c|}{43.71} 
            & 29.89 
            & 52.85 
            & 38.20 
            & 42.17 
            & 25.04 
            & \multicolumn{1}{c|}{35.58} 
            & 52.43$^{\textcolor{red}{\downarrow1.00}}$ \\
            \cmidrule(lr){1-14}

            \multicolumn{14}{l}{\cellcolor[HTML]{EFEFEF}\textbf{\textit{Memory Retrieval Phase}}} \\

            \textit{w/o} Semantic Evaluator 
            & \multicolumn{1}{|c}{52.20} 
            & 66.83 
            & 56.45 
            & 70.92 
            & 32.27 
            & \multicolumn{1}{c|}{42.95} 
            & 29.77 
            & 52.20 
            & 36.50 
            & 40.95 
            & 23.67 
            & \multicolumn{1}{c|}{35.28} 
            & 51.52$^{\textcolor{red}{\downarrow1.91}}$ \\

            \textit{w/o} Counterfactual Inference 
            & \multicolumn{1}{|c}{54.14} 
            & 68.67 
            & 59.15 
            & 72.37 
            & 33.66 
            & \multicolumn{1}{c|}{44.05} 
            & 29.65 
            & 52.31 
            & 37.45 
            & 41.26 
            & 24.44 
            & \multicolumn{1}{c|}{35.39} 
            & 52.34$^{\textcolor{red}{\downarrow1.09}}$ \\
            \bottomrule
        \end{tabular}
    }
\end{table*}

\begin{table*}[th]
    \centering
    \caption{\label{tab:search_hyperparameter_results}
    Search hyperparameter analysis of \ours on \textbf{Qwen3-32B}. We vary one hyperparameter at a time while keeping others fixed to the default setting: $k=5$, $D_{\text{max}}=3$, $m=5$, $\theta_{\text{stable}}=0.8$, $\delta_{QR}=0.5$, and $\delta_{LC}=0.5$. 
    ``\textbf{Rel. Latency}'': the normalized retrieval latency relative to the best-performing default configuration, where $1.00$ corresponds to the default setting and larger values indicate higher retrieval cost. 
    ``\textbf{\#Nodes}'': the average number of memory graph nodes visited during memory retrieval, reflecting the explored search-space size. 
    The best and second best results are highlighted in \textbf{bold} and \underline{underlined}.
    }
    \resizebox{\linewidth}{!}{
    \begin{tabular}{c|cccccc|cccccc|ccc}
        \toprule
        \multirow{3}{*}{\textbf{Hyperparameter Setting}} 
        & \multicolumn{6}{c|}{\textit{Dialogue-Style Memory Benchmarks}} 
        & \multicolumn{6}{c|}{\textit{Long-Form, Document-Style Benchmarks}} 
        & \multirow{3}{*}{\textbf{Avg.}} 
        & \multirow{3}{*}{\textbf{\makecell{Rel.\\Latency}}} 
        & \multirow{3}{*}{\textbf{\#Nodes}} \\
        \cmidrule(lr){2-7} \cmidrule(lr){8-13}
        & \multicolumn{2}{c}{\textbf{LoCoMo}} 
        & \multicolumn{2}{c}{\textbf{LongMemEval}} 
        & \multicolumn{2}{c|}{\textbf{PersonaMem}}
        & \multicolumn{2}{c}{\textbf{HotpotQA}} 
        & \multicolumn{2}{c}{\textbf{2WikiMultihopQA}} 
        & \multicolumn{2}{c|}{\textbf{MuSiQue}}
        &  &  &  \\
        & \textit{EM} & \textit{LJ} & \textit{EM} & \textit{LJ} & \textit{EM} & \textit{LJ} 
        & \textit{EM} & \textit{LJ} & \textit{EM} & \textit{LJ} & \textit{EM} & \textit{LJ} 
        & \textit{LJ} & &  \\
        \midrule
        \rowcolor{gray!15}
        \multicolumn{16}{l}{\textbf{\textit{Beam Width $k$}}} \\
        $k=1$  
        & 50.32 & 64.01 & 55.21 & 68.10 & 30.06 & 40.42 
        & 26.44 & 48.27 & 34.90 & 37.14 & 21.38 & 30.98 
        & 48.15 & 0.42 & 8.6 \\
        $k=3$  
        & 53.21 & 68.15 & 58.73 & 72.18 & 32.91 & 43.96 
        & 29.32 & 52.29 & 37.74 & 41.24 & 24.36 & 34.92 
        & 52.12 & 0.73 & 16.1 \\
        $\mathbf{k=5}$ 
        & \textbf{54.88} & \textbf{69.47} & \textbf{60.14} & \textbf{73.49} & \textbf{34.32} & \textbf{45.05} 
        & \textbf{30.56} & \textbf{53.71} & \textbf{39.07} & \textbf{42.79} & \textbf{25.57} & \textbf{36.04} 
        & \textbf{53.43} & 1.00 & 23.8 \\
        $k=8$  
        & \underline{54.37} & \underline{69.05} & \underline{59.71} & \underline{73.10} & \underline{34.03} & \underline{44.76} 
        & \underline{30.11} & \underline{53.20} & \underline{38.62} & \underline{42.36} & \underline{25.15} & \underline{35.71} 
        & \underline{53.03} & 1.39 & 35.7 \\
        $k=10$ 
        & 53.70 & 68.42 & 59.09 & 72.48 & 33.41 & 44.10 
        & 29.67 & 52.65 & 38.10 & 41.82 & 24.69 & 35.10 
        & 52.43 & 1.65 & 43.9 \\
        \midrule

        \rowcolor{gray!15}
        \multicolumn{16}{l}{\textbf{\textit{Maximum Search Depth $D_{\text{max}}$}}} \\
        $D_{\text{max}}=1$ 
        & 47.90 & 61.02 & 52.32 & 65.87 & 29.78 & 39.50 
        & 24.31 & 45.70 & 31.85 & 35.21 & 19.84 & 29.03 
        & 46.06 & 0.38 & 9.4 \\
        $D_{\text{max}}=2$ 
        & 53.21 & 67.90 & 58.72 & 71.86 & 32.85 & 43.51 
        & 28.94 & 51.86 & 36.85 & 40.08 & 23.90 & 34.38 
        & 51.60 & 0.69 & 17.3 \\
        $\mathbf{D_{\text{max}}=3}$ 
        & \underline{54.88} & \underline{69.47} & \textbf{60.14} & \textbf{73.49} & \underline{34.32} & \underline{45.05} 
        & \underline{30.56} & \underline{53.71} & \underline{39.07} & \underline{42.79} & \textbf{25.57} & \textbf{36.04}  
        & \underline{53.43} & 1.00 & 23.8 \\
        $D_{\text{max}}=4$ 
        & \textbf{55.25} & \textbf{69.84} & \underline{59.58} & \underline{72.96} & \textbf{34.94} & \textbf{45.52} 
        & \textbf{31.12} & \textbf{54.19} & \textbf{39.61} & \textbf{43.12} & \underline{25.09} & \underline{35.66} 
        & \textbf{53.54} & 1.32 & 32.5 \\
        $D_{\text{max}}=5$ 
        & 53.51 & 68.10 & 58.84 & 72.12 & 33.25 & 43.87 
        & 29.45 & 52.48 & 37.92 & 41.50 & 24.50 & 34.95 
        & 52.17 & 1.58 & 40.8 \\
        \midrule

        \rowcolor{gray!15}
        \multicolumn{16}{l}{\textbf{\textit{Candidate Expansion Size $m$}}} \\
        $m=3$  
        & 52.72 & 67.56 & 58.29 & 71.35 & 32.54 & 43.22 
        & 28.55 & 51.10 & 36.48 & 39.85 & 23.42 & 33.77 
        & 51.14 & 0.82 & 18.6 \\
        $\mathbf{m=5}$  
        & \underline{54.88} & \underline{69.47} & \underline{60.14} & \underline{73.49} & \textbf{34.32} & \textbf{45.05} 
        & \textbf{30.56} & \textbf{53.71} & \underline{39.07} & \underline{42.79} & \textbf{25.57} & \textbf{36.04}  
        & \underline{53.43} & 1.00 & 23.8 \\
        $m=10$ 
        & \textbf{55.48} & \textbf{70.12} & \textbf{60.80} & \textbf{74.02} & \underline{34.01} & \underline{44.61} 
        & \underline{30.23} & \underline{53.25} & \textbf{39.71} & \textbf{43.18} & \underline{25.22} & \underline{35.72} 
        & \textbf{53.48} & 1.31 & 33.5 \\
        $m=15$ 
        & 53.65 & 68.31 & 59.11 & 72.41 & 33.42 & 43.97 
        & 29.61 & 52.51 & 38.05 & 41.45 & 24.72 & 35.09 
        & 52.29 & 1.58 & 42.6 \\
        $m=20$ 
        & 52.81 & 67.44 & 58.25 & 71.62 & 32.79 & 43.28 
        & 28.96 & 51.76 & 37.36 & 40.79 & 24.01 & 34.20 
        & 51.52 & 1.82 & 51.3 \\
        \midrule

        \rowcolor{gray!15}
        \multicolumn{16}{l}{\textbf{\textit{Stable Confidence Threshold $\theta_{\text{stable}}$}}} \\
        $\theta_{\text{stable}}=0.6$ 
        & 53.10 & 67.95 & 58.52 & 71.74 & 33.02 & 43.52 
        & 28.82 & 51.36 & 36.92 & 40.17 & 23.83 & 33.81 
        & 51.43 & 1.34 & 35.4 \\
        $\theta_{\text{stable}}=0.7$ 
        & \underline{54.24} & \underline{68.98} & \underline{59.42} & \underline{72.71} & \underline{33.71} & \underline{44.31} 
        & \underline{29.88} & \underline{52.79} & \underline{38.36} & \underline{41.83} & \underline{24.94} & \underline{35.42} 
        & \underline{52.67} & 1.15 & 29.1 \\
        $\mathbf{\theta_{\text{stable}}=0.8}$ 
        & \textbf{54.88} & \textbf{69.47} & \textbf{60.14} & \textbf{73.49} & \textbf{34.32} & \textbf{45.05} 
        & \textbf{30.56} & \textbf{53.71} & \textbf{39.07} & \textbf{42.79} & \textbf{25.57} & \textbf{36.04} 
        & \textbf{53.43} & 1.00 & 23.8 \\
        $\theta_{\text{stable}}=0.9$ 
        & 51.62 & 66.74 & 56.84 & 70.35 & 31.63 & 42.50 
        & 27.56 & 50.11 & 35.89 & 39.07 & 22.83 & 32.68 
        & 50.24 & 0.76 & 15.2 \\
        \midrule

        \rowcolor{gray!15}
        \multicolumn{16}{l}{\textbf{\textit{Counterfactual Inference Thresholds $\delta_{QR}$ and $\delta_{LC}$}}} \\
        $\delta_{QR}=0.3,\delta_{LC}=0.3$ 
        & 53.02 & 67.82 & 58.43 & 71.70 & 32.97 & 43.35 
        & 28.70 & 51.42 & 36.84 & 40.04 & 23.61 & 33.68 
        & 51.34 & 0.91 & 22.1 \\
        $\delta_{QR}=0.4,\delta_{LC}=0.4$ 
        & \underline{54.41} & \underline{69.04} & 59.36 & 72.72 & 33.65 & 44.25 
        & 29.77 & 52.67 & 38.08 & 41.56 & \underline{25.02} & \underline{35.65} 
        & 52.65 & 0.95 & 22.8 \\
        $\mathbf{\delta_{QR}=0.5,\delta_{LC}=0.5}$ 
        & \textbf{54.88} & \textbf{69.47} & \textbf{60.14} & \textbf{73.49} & \textbf{34.32} & \textbf{45.05} 
        & \textbf{30.56} & \textbf{53.71} & \textbf{39.07} & \textbf{42.79} & \textbf{25.57} & \textbf{36.04}  
        & \textbf{53.43} & 1.00 & 23.8 \\
        $\delta_{QR}=0.6,\delta_{LC}=0.6$ 
        & 54.07 & 68.85 & \underline{59.72} & \underline{72.99} & \underline{33.96} & \underline{44.69} 
        & \underline{30.13} & \underline{53.23} & \underline{38.54} & \underline{42.31} & 24.78 & 35.21 
        & \underline{52.88} & 1.16 & 27.9 \\
        $\delta_{QR}=0.7,\delta_{LC}=0.7$ 
        & 53.52 & 68.12 & 58.91 & 71.86 & 33.22 & 43.88 
        & 29.38 & 52.11 & 37.68 & 41.20 & 24.26 & 34.69 
        & 51.98 & 1.38 & 33.4 \\
        \bottomrule
    \end{tabular}
    }
    \vspace{-0.2cm}
\end{table*}

\begin{figure*}[tb]
    \centering
    \includegraphics[width=\textwidth]{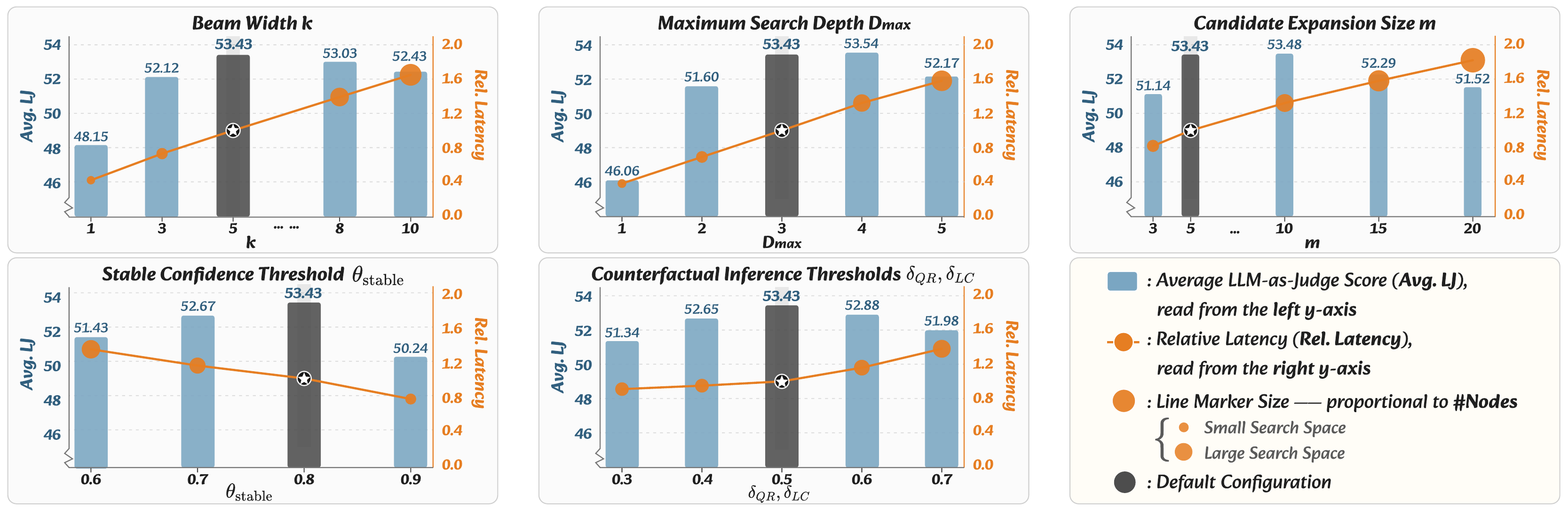}
    \caption{Search hyperparameter statistics of REAL. 
    }
    \label{fig:hyperparameter_tradeoff}
    \vspace{-3mm}
\end{figure*}

\begin{table}[t]
\centering
\caption{\label{tab:resource_overhead}
Resource overhead comparison on the \textbf{Qwen3-32B} model setting. 
``\textbf{Construction Latency}'': the average memory construction latency per dialogue turn. 
``\textbf{Retrieval Latency}'': the average memory retrieval latency per query. 
``\textbf{Answer Latency}'': the average latency of generating the final answer from the retrieved memory context. 
}
\resizebox{\linewidth}{!}{
    \begin{tabular}{lccccc}
        \toprule
        \multirow{2}{*}{\textbf{Method}} 
        & \multicolumn{3}{c}{\textbf{Latency (ms)}}
        & \multirow{2}{*}{\textbf{\#Nodes}} 
        & \multirow{2}{*}{\textbf{Avg. \textit{LJ}}} \\
        \cmidrule(lr){2-4}
        & \textbf{Construction} 
        & \textbf{Retrieval} 
        & \textbf{Answer} & \\ 
        \midrule
        Full Context 
        & -- 
        & -- 
        & 12869.14 
        & -- 
        & 37.13 \\

        Flat Memory Chunk 
        & 248.73 
        & 681.65 
        & 2643.05 
        & -- 
        & 42.47 \\
        
        MemGPT 
        & 1123.69 
        & 1362.48 
        & 2185.38 
        & -- 
        & 44.61 \\
        
        Mem0 
        & 1260.96 
        & 1183.59 
        & 2028.21 
        & -- 
        & 47.83 \\
        
        Vanilla Memory Graph 
        & 1483.67 
        & 1097.54 
        & 2364.52 
        & 41.6 
        & 46.40 \\
        
        Mem0$^g$ 
        & 1762.70 
        & 1620.57 
        & 1857.75 
        & 34.8 
        & 49.56 \\
        
        A-MEM 
        & 1919.72 
        & 1781.05 
        & 1965.81 
        & 31.5 
        & 50.21 \\
        
        \rowcolor{blue!5}
        \textbf{\ours} 
        & 2073.04 
        & 2052.23 
        & 1716.12 
        & 23.8 
        & \textbf{53.43} \\
        \bottomrule
    \end{tabular}
}
\vspace{-0.1cm}
\end{table}

\subsection{Ablation Studies and Discussions}
\label{subsec:ablation}

To provide a detailed understanding of where the performance gains of \ours come from and how the framework behaves under different deployment constraints, we conducted ablation studies on the designed components, the sensitivity analysis of hyperparameters, as well as efficiency assessment, with results shown in \Cref{tab:component_ablation}, \Cref{tab:search_hyperparameter_results} and \Cref{tab:resource_overhead}.

\subsubsection{Component Ablation}
\label{subsubsec:component_ablation}

we perform component ablation on the Qwen3-32B model to quantify the contribution of each major design choice in REAL, including temporal fact modeling, confidence-aware filtering, exploration intent guidance, semantic path scoring, and counterfactual inference.

The results of the component ablation are reported in \Cref{tab:component_ablation}. 
Overall, removing any major component leads to a performance drop, which indicates that the proposed framework benefits from the integration of these core components. 
In particular, removing temporal intervals hurts performance on dialogue-style memory benchmarks by 3.03\%, where user preferences, plans, and relations evolve over time. 
When confidence stratification is removed, answer generation becomes less stable, resulting in an average performance drop of 1.08\% across all datasets. 
Removing the exploration intent label $\iota$ weakens the query-aware nature of the retrieval process, thus leading to a degradation of 2.05\% in overall model performance.
Disabling the semantic evaluator leads to a substantial decline in answer accuracy by 4.44\%, mainly due to poorer memory retrieval quality. 
The performance gap of 2.22\% demonstrates the robustness of memory retrieval by repairing unreliable expansion states through counterfactual inference rather than directly generating unsupported answers.

\subsubsection{Search Hyperparameter Sensitivity}
\label{subsubsec:search_hyper}

We further investigate the sensitivity of hyperparameter $k$, $D_{\text{max}}$, $m$, $\theta_{\text{stable}}$, $\delta_{QR}$, and $\delta_{LC}$ on the Qwen3-32B model to understand the trade-off among accuracy, latency, and search cost, with experimental results summarized in \Cref{tab:search_hyperparameter_results} and \Cref{fig:hyperparameter_tradeoff}.

We experimented with $k$ values of $\left[1,3,5,8,10\right]$ and observed that retrieval performance improved as $k$ increased from 1 to 5, while further increasing $k$ introduced additional semantic evaluation cost and even degraded performance. 
Thus, we set $k=5$ as the default beam width. 
We then varied the maximum search depth $D_{\text{max}}$ in $\left[1,2,3,4,5\right]$ and found that $D_{\text{max}}=3$ was sufficient to capture evidence paths, whereas deeper traversal increased latency and semantic drift. Therefore, we selected $D_{\text{max}}=3$.
For the candidate expansion size $m$, we tested $\left[3,5,10,15,20\right]$. A small $m$ reduced search cost but occasionally missed useful branching evidence, while a large $m$ increased the number of candidate paths requiring semantic evaluation. We found that $m=5$ achieved a favorable trade-off between candidate diversity and computational overhead. 
For confidence-aware retrieval, we varied $\theta_{\text{stable}}$ in $\left[0.6,0.7,0.8,0.9\right]$ and selected $\theta_{\text{stable}}=0.8$, which effectively filtered speculative memories while preserving sufficient evidence coverage. 
Finally, for counterfactual repair, we experimented with $\delta_{QR}$ and $\delta_{LC}$ values in $\left[0.3,0.4,0.5,0.6,0.7\right]$. We observed that lower thresholds under-triggered counterfactual repair, while higher thresholds increased inference cost and introduced more inferred noise. 
Considering the trade-off between repair effectiveness and retrieval efficiency, we set $\delta_{QR}=0.5$ and $\delta_{LC}=0.5$.

\begin{figure*}[t]
    \centering
    \includegraphics[width=\textwidth]{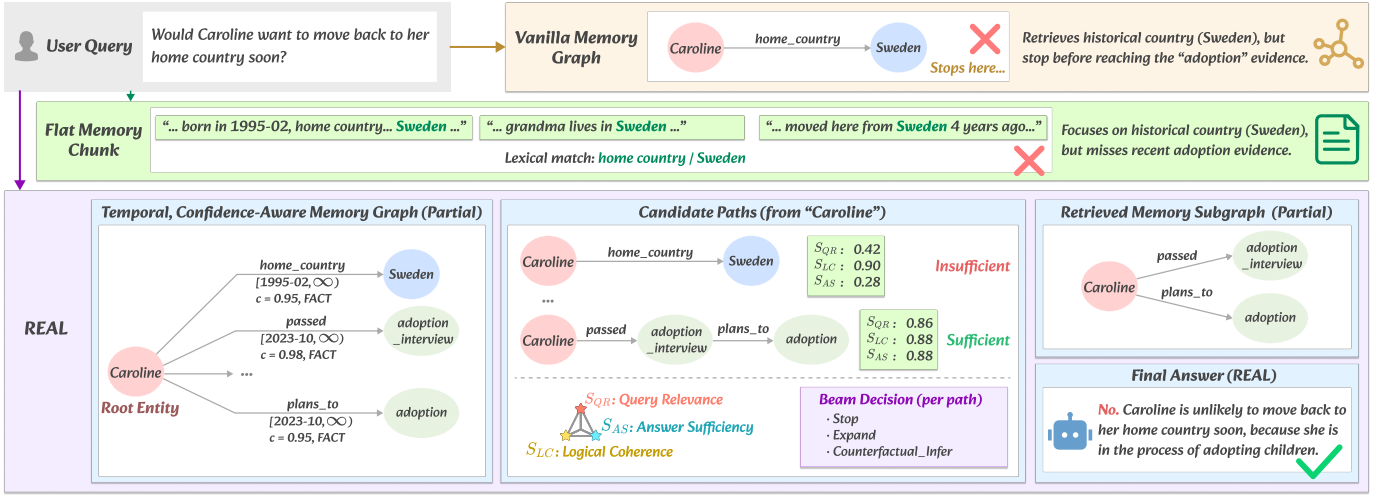}
    \caption{A case study on LoCoMo using DeepSeek-V3 as the backbone model. 
    The phrase ``home country'' is associated with the earlier memory, while answering the query requires the model to further identify Caroline's current life plan.
    }
    \label{fig:case_study}
    \vspace{-2mm}
\end{figure*}

\subsubsection{Efficiency Analysis}
\label{subsubsec:resource_overhead}

We further evaluate the computational resource overhead of \ours across benchmarks on Qwen3-32B deployed with 2$\times$H800 GPUs. 
We report memory construction latency, retrieval latency, answer latency, visited memory nodes (for graph structure memory), and answer accuracy. 
As for \textit{full context}, memory construction, retrieval latency, and visited nodes are not applicable because the entire historical context is directly provided to the answer generator.

As shown in \Cref{tab:resource_overhead}, \ours adds latency during memory construction and retrieval. However, these costs are controlled by hybrid beam search and candidate node filtering. 
Compared to \textit{graph-based} methods, \ours visits the fewest graph nodes and passes a compact memory context to the answer generator while achieving the highest answer accuracy. 
It indicates that \ours makes favorable trade‑offs: the extra computation is effectively converted into superior task performance.

\subsection{Case Studies}
\label{subsec:case_studies}

In \Cref{fig:case_study}, we present a case study from the LoCoMo dataset on the DeepSeek-V3 model to illustrate how \ours differs from typical memory approaches, e.g., \textit{flat memory chunk} and \textit{vanilla memory graph}, in long-term memory management. 
In this case, \ours retrieves a compact memory subgraph that connects \texttt{Caroline}'s current \texttt{adoption process} to her likely future decision, while avoiding distraction from lexically similar but insufficient historical memories.

\section{Related Work}
\label{sec:related_work}
\subsection{Long-Term Memory Management in LLMs}
\label{subsec:llm_memory_related_work}

With the growing deployment of Large Language Models (LLMs) in long-duration applications, such as multi-turn dialogue~\cite{maharana2024evaluating,lu2025datasculpt}, software engineering~\cite{yang2024swe}, and open-world exploration~\cite{shridhar2021alfworld}, 
the demand for persistent, queryable, and evolvable memory has become acute, positioning long-term memory management as a pivotal challenge~\cite{wu2025human,zhang2025memory_survey,kang2025memory}.
A widely adopted approach to memory management in LLM-based systems involves directly incorporating the complete history information as the long-context input at each interaction turn~\cite{beltagy2020longformer,an2024make,fu2024data}.
While such method offers simplicity and effectiveness in low-interaction settings, 
its scalability is constrained by the finite context window of language models.
Concurrently, long contexts often include irrelevant or redundant details that compromise the model's reasoning performance~\cite{shi2023large,liu2024lost,wulongmemeval}.
To overcome the aforementioned limitations, existing solutions can be broadly divided into two categories: 
\textbf{(1) Architectural modifications} embed memory capabilities directly into the model architecture, which include refining position embeddings~\cite{zhao2024length,zheng2024dape}, enhancing attention mechanisms~\cite{liuring2024attention,wang2025m+,kang2025lm2}, and integrating explicit memory modules~\cite{he-etal-2025-hmt}. 
However, these methods depend on access to model-internal states, which is unavailable for closed-sourced or API-based language models.
\textbf{(2) Summarization-oriented approaches} are emerging as a dominant trend, which attempt to manage long-term memory from the perspective of 
context compression, developing techniques to summarize prolonged historical contexts into concise representations~\cite{chevalier2023adapting,jiang2023llmlingua,mu2023learning,xu2023recomp,wang2023augmenting,gutierrez2024hipporag}.

The storage structure and retrieval effectiveness constitute the 
central research emphasis in summarization-oriented memory management approaches. 
Existing methods can be generally classified into two categories: \textit{flat‑text‑based} methods and \textit{graph‑based} methods. 
Flat‑text‑based approaches typically rely on vector embeddings of fixed‑length text chunks stored in a vector database, retrieving the top‑$k$ semantically similar chunks via nearest neighbor search~\cite{zhang2025memory_survey}.
Representative examples include MemGPT~\cite{packer2023memgpt}, 
Mem0~\cite{chhikara2025mem0} 
and RMM~\cite{tan2025prospect}, which continuously update a running summary of the dialogue as new turns arrive. 
However, these strategies struggle to capture the intricate relationships among different pieces of information, and cannot perform multi‑hop reasoning across multiple documents, which impedes the effectiveness of long-term memory organization~\cite{gutierrez2025rag}.

\subsection{Graph-Based Memory Frameworks}
\label{subsec:graph_memory_framework}

To address the above issues, researchers have begun to develop graph‑based strategies by organizing the accumulated contextual data into a graph, 
which serves as a structured memory index, enabling retrieval that goes beyond shallow semantic matching~\cite{liang2025kag,gutierrez2025rag}. 
Typically, Mem0$^g$~\cite{chhikara2025mem0} represents memory as a labeled graph, with entities serving as nodes and relations as edges. 
A-MEM~\cite{xu2025mem} creates interconnected knowledge networks through dynamic indexing and linking of memories. 
HippoRAG~\cite{gutierrez2024hipporag} draws inspiration from the hippocampal memory indexing theory~\cite{teyler1986hippocampal}, and orchestrates LLMs, knowledge graphs, and the Personalized PageRank (PPR) algorithm~\cite{haveliwala2002topic} to perform single‑step, multi‑hop retrieval. 
However, existing graph-based memory methods usually adopt a destructive update paradigm~\cite{tan2026memotime}, and their retrieval procedures are often query-agnostic and operate passively~\cite{ji2026memory}. 
These limitations motivate our temporal, confidence-aware, and reasoning-enhanced graph memory framework.

\section{Conclusion}
\label{sec:conclusion}

In this paper, we proposed REAL, a reasoning-enhanced graph framework for long-term memory management of Large Language Models (LLMs). 
During memory construction, \ours represents historical interactions as a temporal and confidence-aware directed property graph, where atomic facts are annotated with valid-time intervals, confidence scores, and exploration intent labels. 
During memory retrieval, \ours anchors query-relevant root entities, decouples their exploration intents, and performs semantic evaluator-guided hybrid beam search to extract compact memory subgraphs. 
Furthermore, when normal expansion becomes unreliable, counterfactual inference generates alternative traversal hypotheses to recover missing evidence from implicit logical relations.
Extensive experiments demonstrate the effectiveness of \ours over flat-text, graph-based, and existing memory management strategies. 
Overall, our results suggest that treating LLM memory as a structured, temporally evolving, and query-adaptive data management problem is a promising direction for building more reliable long-term LLM systems.

\clearpage

\section*{Acknowledgment}

This work is supported by Fundamental and Interdisciplinary Disciplines Breakthrough Plan of the Ministry of Education of China (JYB2025XDXM113), National Natural Science Foundation of China (92470121, 62402016), National Key R\&D Program of China (2024YFA1014003), Zhongguancun Academy (C20250204, C20250602),  Beijing Major Science and Technology Project (Z251100008125043, Z251100008425023), and High-performance Computing Platform of Peking University.

\section*{AI-Generated Content Acknowledgement}
We acknowledge the use of the generative AI assistants ChatGPT~\cite{brown2020language} and DeepSeek-R1~\cite{guo2025deepseek_r1} in preparing this article. These tools were employed exclusively for language refinement (grammar, word choice, and readability) throughout the manuscript (from \Cref{sec:intro} to \Cref{sec:conclusion}). All core scientific contributions, including the task formalization, system design of our \ours pipeline, experimental strategy, and result analysis, were developed solely by the human authors. The authors carefully reviewed and edited all AI‑suggested modifications to ensure accuracy and alignment with their original work.

\balance
\bibliographystyle{IEEEtran}
\bibliography{icde2027}{}

\end{document}